\documentclass[twoside]{article}

\usepackage[accepted]{aistats2022}
\usepackage[utf8]{inputenc} 
\usepackage[T1]{fontenc}    
\usepackage{hyperref}       
\usepackage{url}            
\usepackage{booktabs}       
\usepackage{amsfonts}       
\usepackage{nicefrac}       
\usepackage{microtype}      
\usepackage{xcolor}         
\usepackage{wrapfig}
\usepackage{dsfont}
\usepackage[ruled]{algorithm2e}

\def\aak#1{{{\color{red}#1}}}
\def\SC{{\rm SC}}
\def\OPT{{\rm OPT}}
\def\CC{{\rm CC}}
\def\UB{{\rm UB}}
\def\SimDP{{\rm SimDP}}
\def\LP{{\rm LP}}
\def\MH{{\rm MH}}
\def\DPI{{\rm DPI}}
\def\DP{{\rm DP}}
\def\sd{{\rm sd}}
\def\Lm{{\rm L}}
\def\tr{{\rm tr}}
\def\te{{\rm te}}
\def\Lh{{\widehat{ L}}}
\def\Lhm{{\widehat{ {\rm L}}}}
\def\Lsep{{{\rm L}^{\rm sep}_{\Zm_{1:N}}}}
\def\Ljnt{{{\rm L}^{\rm joint}_{\Zm_{1:N}}}}
\def\Ber{{\rm Bern}}
\def \Beta{{\rm Beta}}
\def \Bin{{\rm Binomial}}
\def\bbf{{\bf b}}
\def\KL{{\rm KL}}
\def\JS{{\rm JS}}
\def\base{{\rm base}}
\def\meta{{\rm meta}}
\def\Zm{S} 
\def \Pzt{P_{S|\tau}} 
\def \PzTt{P_{S|T=\tau}} 
\def \PzTtt{P_{S|T=T'}} 
\def \Pzti{P_{S|\tau_i}} 
\def \PzT{P_{S|T}} 
\def \PzTi{P_{S|T_i}} 
\def \PZmTi{P_{\Zm_i|\tau}} 
\def \PZMTi{P_{\Zm_i|T}} 
\def \PZmti{P_{\Zm_i|\tau_i}} 
\def \PZMti{P_{\Zm_i|T_i}} 
\def\DPSIED{{\rm DP-SIED}}
\def\DPSI{{\rm DPSI}}
\def\Xbf{\mathbf{X}}
\def\Ybf{\mathbf{Y}}
\def\mset{{\Zbf_{1:N}}}
\def\DPJE{{\rm DPJE}}
\def \Zbf{\mathbf{Z}}
\def \Ztbf{\mathbf{\widetilde{Z}}}
\def\JE{{\rm JE}}
\def\EWRM{{\rm EWRM}}
\def \excess {{\rm excess-risk}}
\def\trte{{\rm sep}}
\def\MER{{\rm MER}}
\def \MEMR{{\rm MEMR}}
\def\LPSI{{\rm LPSI}}
\def \explt{\exp(-\lambda(t|\Hscr(t))\Delta)}
\def\DPSWSID{{\rm DPSW-SID}}
\def\LPSWSID{{\rm LPSW-SID}}
\def\DPSWSIED{{\rm DPSW-SIED}}
\def\DPSWCSI{{\rm DPSW-CSI}}
\def\DPSWJE{{\rm DPSW-JE}}
\def\SW{{\rm SW}}
\def\ind{{\rm ind}}
\def\avg{{\rm avg}}
\def\Ztrain{Z^{\rm m_{tr}}}
\def\Ztest{Z^{\rm m_{te}}}
\def\Zjoint{Z^{\rm m_{jt}}}
\def\Ztr{D^{\rm m}}
\def \Zh{\widehat{Z}}
\def\mtr{m_{\rm tr}}
\def\mjt{m_{\rm jt}}
\def\Zte{D^{\rm m_{te}}}
\def\mte{m_{\rm te}}
\def\Dscrb{\bar{\mathcal{D}}}
\def\Dscrs{\mathcal{D}^{*}}
\def\Dscrt{\widetilde{\mathcal{D}}}
\def \Dscrtr{\mathcal{D}_{\rm{tr}}}
\def \Dscrte{\mathcal{D}_{\rm{te}}}
\def\Sscrm{\pmb{\mathscr{S}}}

\def\independenT#1#2{\mathrel{\rlap{$#1#2$}\mkern2mu{#1#2}}}
\def\csf{\mathsf{c}}
\def\ssf{\mathsf{s}}
\def\xm{\mathrm{x}}
\def \Im{\mathrm{I}}
\def \Tbf{\mathbf{D}}
\def\Ibf{\mathbf{I}}
\def\tfr{\rm{tfr}}
\def\Rbf{\mathbf{R}}
\def\lbf{\mathbf{l}}
\def\Abf{\mathbf{A}}
\def\Sbf{\mathbf{S}}
\def\Zbf{\mathbf{Z}}
\def\Zbft{\mathbf{\widetilde{Z}}}
\def\Iscrm{\mathcal{I}}
\def\Itt{\mathtt{I}}
\def \Sm{\mathrm{S}}
\def\Escr{\mathscr{E}}
\def\lt{\widetilde{l}}
\def\Lh{\widetilde{L}}
\def\thetat{\widetilde{\theta}}
\def\Dscrt{\widetilde{\mathcal{D}}}
\def\lambdat{\widetilde{\lambda}}
\def\Qt{\widetilde{Q}}
\def\Yt{\widetilde{Y}}
\def\qt{\widetilde{q}}
\def\kt{\widetilde{k}}
\def\pt{\widetilde{p}}
\def\zt{\widetilde{z}}
\def\phit{\widetilde{\phi}}
\def\mubf{\mathbf{\mu}}
\def \Nscr{\mathcal{N}}
\def \Wscr{\mathscr{W}}
\def \ge{\mathrm{gen-error}}
\def \DL{ {\rm \Delta} L}
\def \DLm{ {\rm \Delta} {\rm L}}
\def \lh{\widehat{l}}
\def \Lh{\widehat{L}}
\def \mge{\mathrm{metagen-error}}

\usepackage{amsmath, amssymb, bbm, xspace}
\usepackage{epsfig}
\usepackage{longtable}
\usepackage{color}
\usepackage{mathrsfs}
\usepackage{comment}

\usepackage{courier}

\newtheorem{theorem}{Theorem}[section]
\newtheorem{assumption}{Assumption}[section]

\newtheorem{lemma}{Lemma}[section]

\newtheorem{proposition}[theorem]{Proposition}

\newtheorem{corollary}[theorem]{Corollary}

\newtheorem{definition}{Definition}[section]

\def\bkE{{\rm I\kern-.17em E}}
\def\bk1{{\rm 1\kern-.17em l}}
\def\bkD{{\rm I\kern-.17em D}}
\def\bkR{{\rm I\kern-.17em R}}
\def\bkP{{\rm I\kern-.17em P}}

\def\bkZ{{\bf{Z}}}

\def\bkE{{\rm I\kern-.17em E}}
\def\bk1{{\rm 1\kern-.17em l}}
\def\bkD{{\rm I\kern-.17em D}}
\def\bkR{{\rm I\kern-.17em R}}
\def\bkP{{\rm I\kern-.17em P}}

\makeatletter
\newcommand{\pushright}[1]{\ifmeasuring@#1\else\omit\hfill$\displaystyle#1$\fi\ignorespaces}
\newcommand{\pushleft}[1]{\ifmeasuring@#1\else\omit$\displaystyle#1$\hfill\fi\ignorespaces}
\makeatother

\def\bkZ{{\bf{Z}}}
\def\b12{(\beta_1,\beta_2)}

\newcounter{example}
\renewcommand{\theexample}{\thesection.\arabic{example}}

\newcounter{remark}
\renewcommand{\theremark}{\thesection.\arabic{remark}}

\def\Hscr{\mathscr{H}}

\def\Xscr{\mathcal{X}}
\def\Yscr{\mathcal{Y}}

\def\Ebb{\mathbb{E}}
\newlength{\noteWidth}
\long\def\notes#1{\ifinner
{\tiny #1}
\else
\marginpar{\parbox[t]{\noteWidth}{\raggedright\tiny #1}}
\fi\typeout{#1}}

 \def\notes#1{\typeout{read notes: #1}} 

\newcommand{\ie}{i.e.\@\xspace} 
\newcommand{\eg}{e.g.\@\xspace} 

\newcommand{\Real}{\ensuremath{\mathbb{R}}}

\def\OPT{{\rm OPT}}

\def\Ebb{\mathbb{E}}

\def\ind{{\rm ind}}

\def\exp{\mathop{\hbox{\rm exp}}}

\def\spose#1{\hbox to 0pt{#1\hss}}

\def\text #1{\hbox{\quad#1\quad}}

\def\Escr{\mathcal{E}}

\def\nthinsp{\mskip -2   mu}

\def\superstar{^{\raise 0.5pt\hbox{$\nthinsp *$}}}
\def\SUPERSTAR{^{\raise 0.5pt\hbox{$*$}}}

\def\lamstarT {\lambda^{\raise 0.5pt\hbox{$\nthinsp *$}T}}

\def\Ascr{{\cal A}}

\def\Mscr{{\cal M}}

\def\Pscr{{\cal P}}

\def\Uscr{{\cal U}}

\def\Wscr{{\cal W}}
\def\Mscr{{\cal M}}
\def\Nscr{{\cal N}}

\def\Zscr{{\cal Z}}
\def\Xscr{{\cal X}}
\def\Yscr{{\cal Y}}

\def\zbf{{\bf z}}

\def\non{\nonumber}

\let\forallnew\forall
\renewcommand{\forall}{\forallnew\ }
\let\forall\forallnew

		\def\bkE{{\rm I\kern-.17em E}}
		\def\bk1{{\rm 1\kern-.17em l}}
		\def\bkD{{\rm I\kern-.17em D}}
		\def\bkR{{\rm I\kern-.17em R}}
		\def\bkP{{\rm I\kern-.17em P}}
		\def\bkY{{\bf \kern-.17em Y}}
		\def\bkZ{{\bf \kern-.17em Z}}
		\def\bkC{{\bf  \kern-.17em C}}

{\begin{list}{}%
         {\setlength{\leftmargin}{#1}}%
         \item[]%
}
{\end{list}}

		\def\bsp{\begin{split}}
		\def\beq{\begin{eqnarray}}
		\def\bal{\begin{align*}}
		\def\bc{\begin{center}}
		\def\be{\begin{enumerate}}
		\def\bi{\begin{itemize}}
		\def\bs{\begin{small}}
		\def\bS{\begin{slide}}
		\def\ec{\end{center}}
		\def\ee{\end{enumerate}}
		\def\ei{\end{itemize}}
		\def\es{\end{small}}
		\def\eS{\end{slide}}
		\def\eeq{\end{eqnarray}}
		\def\eal{\end{align*}}
		\def\esp{\end{split}}
		\def\qed{ \vrule height7.5pt width7.5pt depth0pt}  

	\def\cp2problem#1#2#3#4{\fbox
		 {\begin{tabular*}{0.9\textwidth}
			{@{}l@{\extracolsep{\fill}}l@{\extracolsep{6pt}}l@{\extracolsep{\fill}}c@{}}
				#1 & & $#4 $ 
			\end{tabular*}}}

		\def\bkE{{\rm I\kern-.17em E}}
		\def\bk1{{\rm 1\kern-.17em l}}
		\def\bkD{{\rm I\kern-.17em D}}
		\def\bkR{{\rm I\kern-.17em R}}
		\def\bkP{{\rm I\kern-.17em P}}
		
		\def\bkZ{{\bf{Z}}}

\newcommand {\beeq}[1]{\begin{equation}\label{#1}}
\newcommand {\eeeq}{\end{equation}}
\newcommand {\bea}{\begin{eqnarray}}
\newcommand {\eea}{\end{eqnarray}}

\def\texitem#1{\par\smallskip\noindent\hangindent 25pt
               \hbox to 25pt {\hss #1 ~}\ignorespaces}

\def\bsp{\begin{split}}
		\def\beq{\begin{eqnarray}}
		\def\bal{\begin{align*}}
		\def\bc{\begin{center}}
		\def\be{\begin{enumerate}}
		\def\bi{\begin{itemize}}
		\def\bs{\begin{small}}
		\def\bS{\begin{slide}}
		\def\ec{\end{center}}
		\def\ee{\end{enumerate}}
		\def\ei{\end{itemize}}
		\def\es{\end{small}}
		\def\eS{\end{slide}}
		\def\eeq{\end{eqnarray}}
		\def\eal{\end{align*}}
		\def\esp{\end{split}}
		\def\qed{ \vrule height7.5pt width7.5pt depth0pt}  

\usepackage{amsmath, amssymb, xspace}
\usepackage{epsfig}
\usepackage{longtable}
\usepackage{color}
\usepackage{mathrsfs}
\usepackage{subfig}
\newenvironment{proof}[1][]{{\noindent \textit{ Proof}: }}{\hfill \qed \vspace{3pt}\\ }

\def\Nscr{{\cal N}}

\begin{document}

%

%

\twocolumn[

\aistatstitle{Information-Theoretic Analysis of Epistemic Uncertainty in  Bayesian Meta-learning }

\aistatsauthor{ Sharu Theresa Jose \And Sangwoo Park \And  Osvaldo Simeone }

\aistatsaddress{Department of Engineering, King's College London, London, UK, WC2R 2LS \\King’s Communications, Learning and Information Processing (KCLIP) Lab \\\texttt{\{sharu.jose, sangwoo.park, simeone.osvaldo\}@kcl.ac.uk}}]

\begin{abstract}
 The overall predictive uncertainty of a trained predictor can be decomposed into separate contributions due to epistemic and aleatoric uncertainty. Under a Bayesian formulation, assuming a well-specified model, the two contributions can be exactly expressed (for the log-loss) or bounded (for more general losses) in terms of information-theoretic quantities \cite{xu2020minimum}. This paper addresses the study of epistemic uncertainty within an information-theoretic framework in the broader setting of Bayesian meta-learning. A general hierarchical Bayesian model is assumed in which hyperparameters determine the per-task priors of the model parameters. Exact characterizations (for the log-loss) and bounds (for more general losses) are derived for the epistemic uncertainty – quantified by the minimum excess meta-risk (MEMR) – of optimal meta-learning rules. This characterization is leveraged to bring insights into the dependence of the epistemic uncertainty on the number of tasks and on the amount of per-task training data. Experiments are presented that use the proposed information-theoretic bounds, evaluated via neural mutual information estimators, to compare the performance of conventional learning and meta-learning as the number of meta-learning tasks increases.
\end{abstract}
\section{Introduction}
\begin{figure}[t]
 \begin{center}
   \includegraphics[scale=0.34,trim=2.6in 1.9in 1.1in 0.55in,clip=true]{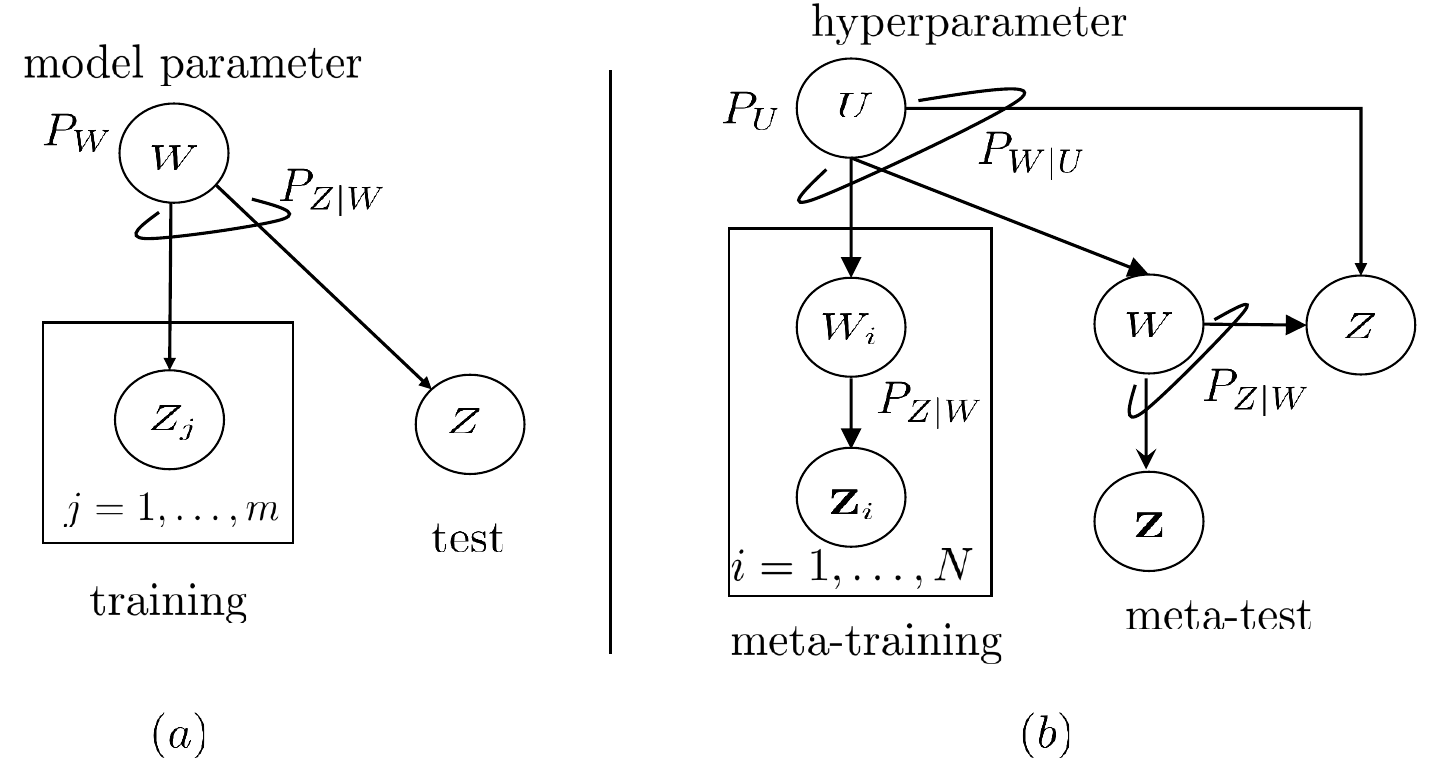} 
   \caption{A graphical model representation of the joint distribution of the relevant quantities for : $(a)$ conventional Bayesian learning; and $(b)$ Bayesian meta-learning.} \label{fig:bayesianML}
   \end{center}
  \end{figure}  
\textbf{Bayesian learning and epistemic uncertainty.} Bayesian machine learning is well understood to have important advantages in terms of uncertainty quantification, model selection, and out-of-distribution detection \cite{mackay2003information}, \cite{wilson2020case}. Bayesian learning assumes the probabilistic model illustrated in Figure~\ref{fig:bayesianML}$(a)$, in which the training data $\Zbf$ is generated in an i.i.d. manner given a model parameter $W$ that is considered to be a random variable endowed with a prior distribution $P_W$.  Assuming that the model is well specified, the overall uncertainty of the optimal predictor for a test target variable $Y$ given input $X$, when measured by the log-loss,  is given by the conditional entropy $H(Y|X,\Zbf)$. This can be decomposed as \cite{xu2020minimum}
\begin{align}
    H&(Y|X,\Zbf)\nonumber\\=&\underbrace{H(Y|X,W)}_{\mbox{aleatoric uncertainty}}+\underbrace{I(Y;W|X,\Zbf),}_{\mbox{epistemic uncertainty}}
\end{align} where the conditional entropy $H(Y|X,W)$ quantifies the \textbf{aleatoric} \textbf{uncertainty} in the prediction, while the conditional mutual information (MI) $I(Y;W|X,\Zbf)$ accounts for the \textbf{epistemic uncertainty}.
The {aleatoric uncertainty} captures the inherent randomness in the data generation process, and is independent of the amount of the available data; while the {epistemic uncertainty}, also known as the \textbf{minimum excess risk} (\textbf{MER}), is caused by limitations in the availability of data, and vanishes as more training data is processed. In this work, we aim at extending this decomposition, and related analysis, from conventional Bayesian learning to Bayesian meta-learning \cite{kim2018bayesian}, \cite{grant2018recasting}, \cite{ravi2019amortized}. 
 
\textbf{Bayesian meta-learning.} In conventional Bayesian learning, the prior $P_W$ of the model parameter is fixed a priori based on knowledge about the problem or tractability. A choice of the prior that matches the data generation mechanism can reduce the amount of data required to meet accuracy requirement. Bayesian meta-learning aims to automatically infer the prior $P_W$ by observing a finite number $N$ of “related” tasks, so that the predictive performance on a new, previously unseen task, in the same class can be improved \cite{kim2018bayesian}, \cite{grant2018recasting}, \cite{ravi2019amortized}. The shared statistical properties of a class of tasks can be modelled using the \textbf{hierarchical Bayesian model} show in Figure~\ref{fig:bayesianML}$(b)$. In it, a \textbf{latent hyperpameter} $U$ determines the prior of all tasks via the conditional distribution as $P_{W|U}$. Assuming a well-specified model, knowing the hyperparameter $U$ hence yields the correct prior $P_{W|U}$ for the new task. The hyperparameter $U$ itself is assumed to be random, and endowed with a \textbf{hyperprior} distribution $P_U$.  
 
\textbf{Contributions.} This work aims at developing, analyzing, and evaluating information-theoretic characterizations of the epistemic  uncertainty associated with a meta-learning predictor. The meta-learner has access to limited labelled data from a set of related tasks, known as \textbf{meta-learning tasks}, as well as from the new task of interest, known as \textbf{meta-test task}. Intuitively, meta-learning data can help reduce the epistemic uncertainty associated with the hyperparameter $U$, while meta-test task data is important to further reduce the epistemic uncertainty at the level of the per-task model parameter $W$. With this in mind, the main contributions of this work are as follows: 

1.	We first develop an exact information-theoretic characterization of the overall epistemic uncertainty for an optimal Bayesian meta-learner for the \textbf{log-loss}.
The characterization is given in terms of the \textbf{minimum excess meta-risk (MEMR)}, which generalize the notion of MER to Bayesian meta-learning. The bound reveals that, under suitable assumptions, the first contribution to epistemic uncertainty – due to the hyperparameter $U$ – scales as  $O(d \log(N) /N)$, where $N$ is the number of meta-training tasks for fixed  number of per-task data samples $m$; while  the contribution due to model parameter uncertainty scales as $O(d \log(m)/m)$. 

2.	We evaluate the derived information-theoretic bounds for the problem of meta-learning the priors of a Bayesian neural network (BNN) by leveraging mutual information neural estimation (MINE) \cite{belghazi2018mutual}, \cite{mukherjee2020ccmi}, and compare the performance of conventional learning and meta-learning as the number of meta-training tasks increses.

Additional material including a generalization to a broader class of loss functions and an information-theoretic comparison of Bayesian learning and meta-learning can be found in the supplementary file.
\section{Related Work}
\paragraph{Information-theoretic generalization analysis.} The works \cite{russo2016controlling}, \cite{xu2017information} have shown that the \textbf{generalization error} of conventional learning algorithms in the frequentist setting
can be upper bounded in terms of the MI $I(W;\Zbf)$ between the input training set and the output model parameter. This metric captures the \textbf{sensitivity} of the learning mechanism to the input training set. Various refinements of these MI-based bounds have been studied since by \cite{bu2019tightening}, \cite{negrea2019information}, \cite{steinke2020reasoning}, among others.

Moving from the frequentist to Bayesian learning, the recent work in \cite{xu2020minimum} introduces an information-theoretic analysis of the MER in Bayesian learning.
The MER for a general class of loss functions, including log-loss and bounded loss, is shown to be upper bounded via functions of the ratio $I(W;\Zbf)/m$ where $m$ is the number of data samples. Under appropriate regularity assumptions on the model, this upper bound is shown to vanish in the limit as $m \rightarrow \infty$.

\textbf{Generalization analysis of meta-learning.} Originating in the work by \cite{schmidhuber1987evolutionary} and \cite{thrun1998learning}, meta-learning has been extensively studied in recent years both in terms of algorithm design \cite{finn2017model}, \cite{nichol2018first} and of analytical studies on the \textbf{meta-generalization error} within a frequentist setting \cite{pentina2014pac}, \cite{amit2018meta}, \cite{rothfuss2020pacoh} , \cite{jose2021entropy}. While the works in \cite{pentina2014pac}, \cite{amit2018meta}, \cite{rothfuss2020pacoh} obtain high-probability PAC-Bayesian bounds on the meta-generalization error with respect to the meta-training data, reference \cite{jose2021entropy} presents  information-theoretic bounds on the average meta-generalization error, thereby extending the work of \cite{xu2017information} for conventional learning to meta-learning. 
Refinements and extensions to these bounds have been studied in \cite{rezazadeh2020conditional}, \cite{jose2021tasksimilarity}.

\textbf{Bayesian meta-learning.} Most activity on Bayesian meta-learning has modelled the hyperparameter $U$ in Fig. 1(b) as deterministic and only captured epistemic uncertainty related to the model parameter $W$. The approach is akin to empirical Bayes \cite{grant2018recasting}, and it has been investigated in \cite{finn2018probabilistic}, \cite{kim2018bayesian}, \cite{ravi2019amortized}, \cite{gordon2018meta},  \cite{nguyen2020uncertainty}. Notably, reference \cite{kim2018bayesian} proposed the use of Stein Variational Gradient Descent (SVGD) \cite{liu2016stein} to carry out non-parametric variational inference (VI) for $W$. SVGD is more flexible and can be effective than standard parametric VI methods based on Gaussian distributions \cite{liu2018riemannian}. Fully Bayesian meta-learning methods were derived in \cite{rothfuss2020pacoh}, \cite{amit2018meta} from a PAC Bayes perspective by using parametric VI with Gaussian models. In this paper, we are not concerned with introducing new approximate meta-learning algorithms, but rather to evaluate the generalization performance of exact Bayesian meta-learning.
\section{Problem Setting}
In this section, we first review the setting studied in \cite{xu2020minimum} for conventional Bayesian learning along  with the key definition of Minimum Excess Risk (MER). Then, we generalize the framework to the Bayesian meta-learning setup introduced and analyzed in this paper. Central to our analysis is the Minimum Excess Meta-Risk (MEMR) metric, which extends the MER to meta-learning. We adopt standard notations for information-theoretic quantities such as (conditional) entropy and (conditional) MI as defined in \cite{cover06elements}.
\subsection{Conventional Bayesian Learning}\label{sec:conventional_learning}
In supervised learning, each data point $Z=(X,Y) \in \Zscr$ consists of a tuple of input feature vector $X \in \Xscr$ and target variable $Y \in \Yscr$, which is drawn from an unknown population distribution. The learner observes a training data set $\mathbf{Z}=(Z_1,\hdots,Z_m)$, of $m$ samples, $Z_i=(X_i,Y_i)$ for $i=1,\hdots, m$, that are generated i.i.d. according to the underlying unknown distribution, and uses it to predict the label of a test feature input $X$ drawn independently from the training set $\Zbf$ from the same distribution. Considering a \textbf{parametric generative model}, we assume that the unknown population distribution  belongs to a model class $\Mscr=\{P_{Z|w}: w \in \Wscr\}$ parametrized by a model parameter $w$ in the set $\Wscr$. This implies that the model class is \textbf{well-specified} \cite{knoblauch2019generalized}. 

In conventional Bayesian learning, the model parameter $W$ is treated as a latent random vector, and is endowed with a prior distribution $P_W$. Conditioned on the model parameter $W$, the data samples are drawn i.i.d. from the model $P_{Z|W}$. Consequently, the joint distribution of model parameter $W$, training set $\Zbf$, and test sample $Z=(X,Y)$ is given as the product
\begin{align}
P_{W,\Zbf,Z}=P_W \otimes P^{\otimes m}_{Z|W} \otimes P_{Z|W}, \label{eq:jointdistribution}
\end{align} 
which factorizes according to the Bayesian network illustrated in Figure~\ref{fig:bayesianML}$(a)$.

 Let $\Ascr$ denote an action space and $\ell:\Yscr \times \Ascr \rightarrow \Real$ denote a loss function. The loss accrued by action $a \in \Ascr$ on target variable $y\in \Yscr$ is measured by the loss function $\ell(y,a)$.
Under the generative model \eqref{eq:jointdistribution}, the Bayesian learning problem is to infer a \textbf{decision rule},
$
\Psi_{\base}: \Zscr^m \times \Xscr \rightarrow \Ascr $, 
mapping the input training data $\Zbf \in \Zscr^m$ and test input $X \in \Xscr$ to an action $a \in \Ascr$, that minimizes the expected loss $\Ebb_{P_{Y,X,\Zbf}}[\ell(Y,\Psi_{\base}(X,\Zbf))]$, where $P_{Y,X,\Zbf}$ is the marginal of \eqref{eq:jointdistribution} over $Y,X,$ and $\Zbf$.
\begin{definition}\cite{xu2020minimum}
The \textbf{Bayesian risk} for a loss function $\ell:\Yscr \times \Ascr \rightarrow \Real$ is  the minimum expected loss across all possible choices of the decision rule, \ie,
\begin{align}
R_{\ell}&(Y|X,\Zbf)\nonumber\\&:=\min_{\Psi_{\base}:\Zscr^m \times \Xscr \rightarrow \Ascr} \Ebb_{P_{Y,X,\Zbf}}\Bigl[\ell(Y,\Psi_{\base}(\Zbf,X))\Bigr]. \label{eq:Bayesianrisk}
\end{align}
\end{definition}
The Bayesian risk is lower bounded by the expected loss obtained by an ideal decision rule, $\Phi_{\base}:\Wscr \times \Xscr \rightarrow \Ascr$, that has access to the true model parameter $W$ generating the test sample $Z=(X,Y) \sim P_{Z|W}$.
\begin{definition}
The \textbf{genie-aided Bayesian risk} is defined as
\begin{align}
R_{\ell}&(Y|X,W)\nonumber \\&:=\min_{\Phi_{\base}:\Wscr \times \Xscr \rightarrow \Ascr} \Ebb_{P_{Y,X,W}}\Bigl[\ell(Y,\Phi_{\base}(W,X))\Bigr]. \label{eq:genieaidedrisk}
\end{align}
The difference between the Bayesian risk \eqref{eq:Bayesianrisk} and the genie-aided risk \eqref{eq:genieaidedrisk} is the \textbf{minimum excess risk (MER)},
\begin{align}
\MER_{\ell}:=R_{\ell}(Y|X,\Zbf)-R_{\ell}(Y|X,W). \label{eq:MER_definition}
\end{align}
\end{definition}

The MER defined in \eqref{eq:MER_definition} satisfies the following properties (\cite{xu2020minimum}): $(i)$ $\MER_{\ell} \geq 0$; and $(ii)$ it is non-increasing with respect to the number of data samples $m$. Importantly, by \eqref{eq:MER_definition}, the minimum predictive uncertainty, $R_{\ell}(Y|X,\Zbf)$, can be written as the sum $R_{\ell}(Y|X,W)+\MER_{\ell}$ of genie-aided Bayesian risk and MER. The first term quantifies  the \textbf{aleatoric uncertainty} resulting from the inherent presence of randomness in the data generation process; while the MER quantifies the \textbf{epistemic uncertainty} due to availability of insufficient data to identify the model parameter $W$. 
\subsection{Bayesian Meta-Learning} 
In conventional Bayesian learning, the prior distribution $P_W$ on the model parameters is conventionally chosen based on prior knowledge about the problem. In contrast, in Bayesian meta-learning, this selection is \textit{data-driven} and \textit{automated}. Specifically, by observing data from  a number $N$ of tasks with shared statistical characteristics, meta-learning aims at inferring a suitable prior $P_W$, to be used on a new, a priori unknown task. As we detail next, the statistical relationship among different tasks is accounted for via a hierarchical Bayesian model that includes a global latent hyperparameter $U \in \Uscr$ \cite{gordon2018decision}. 
As illustrated in Figure~\ref{fig:bayesianML}$(b)$, the meta-learner is given data from $N$ \textbf{meta-training tasks}. Data for each task $i$ is drawn from the distribution $P_{Z|W=W_i}$ with the task-specific model parameter $W_i$. In particular, conditioned on model parameter $W_i$, the training data samples of 
 each $i$th task, $\Zbf_i=(Z_1^{i},\hdots,Z_m^{i})$, are i.i.d. and drawn from the data distribution $P_{Z|W=W_i}$. 
The model parameter $W_i$ of each task $i$ is drawn from a shared prior distribution $P_{W|U}$, parameterized by a common hyperparameter $U$. Both the model parameter $W_i$ and hyperparameter $U$ are assumed to be latent random variables, with joint distribution factorizing  as $P_U \otimes P_{W|U}$, with $P_U$ denoting the \textbf{hyper-prior} distribution. The parameterized prior $P_{W|U}$ is assumed to be the same for all tasks, and the statistical relationship of the observed tasks is captured through the hyperparameter $U$. 


  The \textbf{meta-training 
set} $\Zbf_{1:N}=(\Zbf_1, \hdots, \Zbf_{N})$ includes the data sets from the $N$ meta-training tasks. The goal is to use this data to reduce the expected loss measured on a  \textbf{meta-test task}. The latter is a priori unknown, and is modelled as being generated by drawing an independent model parameter $W \sim P_{W|U}$ for the given hyperparameter $U$ that is shared with the meta-training data. This model parameter underlies the generation of the \textbf{meta-test training data} $\Zbf \sim P^{\otimes m}_{Z|W}$, and an independently generated \textbf{meta-test test data} $Z=(X,Y) \sim P_{Z|W}$. 

To summarize, as shown in Figure~\ref{fig:bayesianML}$(b)$, the joint distribution of  global parameter $U$, model parameters $W_{1:N}=(W_1,\hdots,W_N)$ for the meta-training tasks,  meta-training data set $\Zbf_{1:N}$, training data $\Zbf$ and test data $Z$ of the meta-test task with model parameter $W$, is given as
\begin{align}
&P_{U,W_{1:N},\Zbf_{1:N},W,\Zbf,Z}\nonumber \\&=P_U \otimes \underbrace{\biggl(P_{W|U}\otimes P_{\Zbf|W}\biggr)^{\otimes N}}_{\mbox{meta-training}} \otimes \underbrace{P_{W|U}\otimes P_{\Zbf|W} \otimes P_{Z|W}}_{\mbox{meta-testing}}.\label{eq:metalearning_distribution}
\end{align}

 The \textbf{meta-learning decision rule} is defined as a mapping
$
\Psi_{\meta}: \Zscr^{Nm} \times \Zscr^m \times \Xscr \rightarrow \Ascr
$
from observed meta-training set $\Zbf_{1:N} \in \Zscr^{Nm}$, training set $\Zbf \in \Zscr^m$  and  test feature input $X \in \Xscr$ of the meta-test task to the action space $\Ascr$. In words, the meta-learning rule $\Psi_{\meta}$ leverages meta-training data, along with the training data for the meta-test task, to predict the target variable $Y$ of the test sample $(X,Y)$ for the meta-test task.
Under the meta-learning generative model in \eqref{eq:metalearning_distribution}, the Bayesian meta-learning problem is to infer a decision rule
$
\Psi_{\meta}(\mset,\Zbf,X)
$
so as to minimize the expected loss $\Ebb_{P_{\mset,\Zbf,X,Y}}[\ell(Y,\Psi_{\meta}(\Zbf_{1:N},\Zbf,X))]$,
where $P_{\mset,\Zbf,X,Y}$ is the marginal of the joint distribution \eqref{eq:metalearning_distribution} over $(\mset,\Zbf,X,Y)$. Accordingly, we  have the following definitions.
\begin{definition}
For a given loss function $\ell:\Yscr \times \Ascr \rightarrow \Real$, the \textbf{Bayesian meta-risk} is the minimum expected loss across all possible choices of the meta-learning decision rule
\begin{align}
R_{\ell}(Y&|X,\Zbf_{1:N},\Zbf)\nonumber\\:=&\min_{\Psi_{\meta}:\Zscr^{Nm} \times \Zscr^m \times \Xscr \rightarrow \Ascr} \nonumber \\ &\quad\quad \Ebb_{P_{\mset,\Zbf,X,Y}}\Bigl[\ell(Y,\Psi_{\meta}(\Zbf_{1:N},\Zbf,X))\Bigr]. \label{eq:Bayesianrisk_meta}
\end{align}
\end{definition}

The Bayesian meta-risk reduces to the conventional Bayesian risk \eqref{eq:Bayesianrisk} when the per-task prior distribution $P_{W|U}$ does not depend on the hyperparameter $U$, and no meta-training set is observed \ie, 
\begin{align}
   \mset=\emptyset, \quad \mbox{and} \quad  P_{W|U}=P_W. \label{eq:conditions}
\end{align} The Bayesian meta-risk is lower bounded by the expected loss obtained by a genie-aided decision rule, $\Phi_{\meta}:\Uscr \times \Wscr \times \Xscr \rightarrow \Ascr$  that has access to the true shared hyperparameter $U$ and the true model parameter $W$ of the test task.
\begin{definition}
The \textbf{genie-aided Bayesian meta-risk} is defined as
\begin{align}
&R_{\ell}(Y|X,W,U)\nonumber \\&:=\min_{\Phi_{\meta}:U \times W \times \Xscr \rightarrow \Ascr} \Ebb_{P_{U,W,X,Y}}[\ell(Y,\Phi_{\meta}(U,W,X))]. \label{eq:genie_metarisk}
\end{align}
The difference between the Bayesian meta-risk in \eqref{eq:Bayesianrisk_meta} and the genie-aided Bayesian meta-risk in \eqref{eq:genie_metarisk} is the \textbf{minimum excess meta risk (MEMR)}, \ie,
\begin{align}
\MEMR_{\ell}:=R_{\ell}(Y|X,\Zbf_{1:N},\Zbf)-R_{\ell}(Y|X,W,U). \label{eq:MEMR_def}
\end{align}
\end{definition}

The MEMR reduces to the MER when condition \eqref{eq:conditions} holds.
\section{Information-Theoretic Analysis of the MEMR}
In this section, we first provide some general properties of the MEMR. Then, we analyze the MEMR with log-loss as the loss function, and obtain information-theoretic upper bounds that explicitly reveal the dependence of the MEMR on the number of meta-training tasks and per-task data samples. The detailed proofs of all the results can be found in the supplementary material.
\subsection{Exact Analysis of the MEMR}
Generalizing the properties of the MER reviewed in Section~\ref{sec:conventional_learning} and proved in \cite{xu2020minimum}, the MEMR can be shown to satisfy the following properties.
\begin{lemma}\label{lem:properties_MEMR}
The minimum excess meta-risk $\MEMR_{\ell}$ is non-negative, \ie,  $\MEMR_{\ell} \geq 0$ and it is non-increasing with respect to the number of tasks, $N$, and to number of data samples per task, $m$. 
\end{lemma}

We now evaluate the MEMR explicitly in terms of information-theoretic metrics when the loss function $\ell(\cdot,\cdot)$ is the log-loss. To this end,
consider the action space $\Ascr$ to be the space of all probability distributions $q(\cdot)$ on $\Yscr$. We assume that all necessary measurability conditions are satisfied \cite{xu2020minimum}. The log-loss accrued by distribution $q(\cdot)$ on a given target $y$ is defined as 
$
\ell(y,q)=-\log q(y).$

\paragraph{MER for conventional Bayesian learning.} For reference, we first review a result from \cite{xu2020minimum} that expresses the MER for conventional Bayesian learning in terms of a conditional MI, and bounds it as a function of a scaled MI.
\begin{lemma}\cite{xu2020minimum}
The minimum excess risk \eqref{eq:MER_definition} for the log-loss satisfies
\begin{align}
    \MER_{\log}&=I(Y;W|X,\Zbf)=H(Y|X,\Zbf)-H(Y|X,W) \label{eq:MER_xu}  \\
    &\leq \frac{I(W;\Zbf)}{m} \label{eq:MER_ub}. 
\end{align}
\end{lemma}

In \eqref{eq:MER_xu}, the conditional entropy $H(Y|X,\Zbf)$ captures the \textbf{overall predictive uncertainty} of the target $Y$ when tested on the feature input $X$ using the training data $\Zbf$, while the term $H(Y|X,W)$ accounts for the \textbf{aleatoric uncertainty}. The latter results from the inherent randomness in the observations, which applies even when true model parameter $W$ is known. The  difference between the two yield the conditional mutual information $I(Y;W|X,\Zbf)$, which captures the \textbf{epistemic uncertainty} in predicting $Y$. In \eqref{eq:MER_ub}, the MER is upper bounded by the term that depends on the MI $I(W;\Zbf)$ between the model parameter and the training data.  This term captures the ``\textbf{sensitivity}'' of the trained model parameter $W$ on the training data $\Zbf$ (see Sec. 2), in the sense that it quantifies the dependence of the trained model parameter $W$ on $\Zbf$ \cite{raginsky2016information}.

\paragraph{MEMR for Bayesian meta-learning.} Our first main result is the generalization of the information-theoretic characterization \eqref{eq:MER_xu} to Bayesian meta-learning.

\begin{proposition}\label{prop:MEMR_log}
The minimum excess meta-risk \eqref{eq:MEMR_def} for the log-loss is given as
\begin{align}
    \MEMR_{\log}&=I(Y;W|X,\Zbf,\Zbf_{1:N}) \non \\
    &=H(Y|X,\Zbf,\mset)-H(Y|X,W). \label{eq:MEMR_log}
\end{align}
\end{proposition}
The proof can be found in the supplementary materials.

In \eqref{eq:MEMR_log}, the conditional entropy $H(Y|X,W)$ captures the \textbf{aleatoric uncertainty} in predicting $Y$, which applies even when the true  model parameter $W$ is known. In contrast, the term $H(Y|X,\Zbf,\mset)$ captures the average predictive uncertainty of the optimal meta-learning decision rule.  As such, the conditional MI $I(Y;W|X,\Zbf,\Zbf_{1:N})$
captures the \textbf{epistemic uncertainty}. This uncertainty results from the availability of limited meta-training tasks and meta-test training data, which causes the true model parameter $W$ and true hyperparameter $U$ to be inaccurately estimated.

\paragraph{Dependence of MEMR on $N$ and $m$.} We now decouple the contributions of hyperparameter-level and per-task-level uncertainties by developing an information-theoretic upper bound on the MEMR \eqref{eq:MEMR_log}. The bound will be used to relate the MEMR to the number of meta-training tasks, $N$, and to the number of samples of the meta-test training set, $m$.
\begin{theorem}\label{thm:loglossUB}
The following upper bounds on the MEMR hold under the log-loss,
\begin{align}
\MEMR_{\log}&\leq \frac{I(W,U;\Zbf|\Zbf_{1:N})}{m} \label{eq:MER_ub1}\\&\leq \frac{I(U;\Zbf_{1:N})}{Nm}+\frac{I(W;\Zbf|U)}{m}=: \MEMR^{\UB}_{\log}, \label{eq:MER_upperbound}
\end{align}where the inequality \eqref{eq:MER_upperbound} holds for $N\geq 1$.
\end{theorem}

The upper bound $\MEMR^{\UB}_{\log}$ \eqref{eq:MER_upperbound} on the MEMR for the log-loss is the sum  of two contributions.
The first term captures the \textbf{sensitivity} of the hyperparameter $U$ on the meta-training set $\mset$.
The second term corresponds to the average \textbf{sensitivity} of the model parameter $W$ on the meta-test task training data $\Zbf$ assuming that the hyperparameter $U$ is known.  The result \eqref{eq:MER_upperbound} shows that the epistemic uncertainty $I(Y;W|X,\Zbf,\mset)$, which applies to the domain of the target variable $Y$,  is upper bounded by the sum of two contributions that pertain the uncertainty levels in the spaces of  hyperparameter  and model parameter, respectively.

The additive dependence of the upper bound \eqref{eq:MER_upperbound} on two mutual information terms, one at the hyperparameter level and other at the per-task model parameter level, bears resemblance to the information-theoretic bounds on the generalization error of frequentist meta-learning problems obtained in \cite{jose2021entropy}, \cite{jose2021tasksimilarity}. However, the two types of bounds are conceptually different. In fact, the generalization error bounds in \cite{jose2021entropy,jose2021tasksimilarity} quantify the error in approximating the meta-population loss with an empirical meta-training loss for an arbitrary stochastic learning algorithm. In contrast, the MEMR metric captures the excess prediction risk obtained by the posterior distribution under the assumption of a well-specified model. As a result, while the information-theoretic upper bounds in  \cite{jose2021entropy,jose2021tasksimilarity} can be used to re-derive the regularized training loss objectives in \cite{rothfuss2020pacoh}, the upper bound \eqref{eq:MER_upperbound} cannot play this role.
 


\textbf{A generalization of Theorem~\ref{thm:loglossUB} to any loss function } and a \textbf{comparison between meta-learning and conventional learning} in terms of predictive accuracy is available in the supplementary materials.

\paragraph{Asymptotic analysis of the MEMR.} The first term in \eqref{eq:MER_upperbound} is a function of $(N,m)$ and the second is of $m$, obscuring the scaling of the MEMR with $(N,m)$. 
To investigate this point, we now study the asymptotic behavior of the above two terms in \eqref{eq:MER_upperbound}.
\begin{lemma} \label{lem:convergencerates}
Let $W \in \Wscr$ and $U \in \Uscr$ be $d$-dimensional vectors taking values in compact subsets $\Wscr,\Uscr \subset  \Real^d$ respectively. Assume that the data distribution $P_{Z|W}(\cdot|w)$ is smooth in $w \in \Wscr$, and that the distribution $P_{\Zbf|U}(\cdot|u)$ is smooth in $u \in \Uscr$. Then, under additional technical conditions (included in supplementary material), we have that for fixed $m$, as $N \rightarrow \infty$,
\begin{align}
   I(U;\Zbf_{1:N})&=\frac{d}{2}\log \Bigl(\frac{N}{2\pi e}\Bigr) +H(U)\non \\&+\Ebb_{P_{U}} \Bigl[ \log |J_{\Zbf|U}(U)| \Bigr]+o(1), \label{eq:asymptotic-1}
\end{align}
and as $m\rightarrow \infty$, we have
\begin{align}
 I(W;\Zbf|U) &=\frac{d}{2}\log \Bigl(\frac{m}{2\pi e}\Bigr)+H(W|U)\non \\&+\Ebb_{ P_{W,U}} \Bigl[ \log |J_{Z|W}(W)| \Bigr]+o(1), \label{eq:asymptotic-2}
\end{align}
where $H(\cdot)$ denotes the differential entropy of the argument random variable and $J_{A|B}(B)$ is the Fisher information matrix (FIM) about $B$ contained in $A$ with respect to conditional distribution $P_{A|B}$, whose $(j,k)$th entry is
\begin{align}
      [J_{A|B}(B)]_{j,k}=\biggl[ \frac{\partial^2}{\partial B_j' \partial B_k'} D_{\KL}(P_{A|B}||P_{A|B'}) \biggr|_{B'=B}\biggr].
      \end{align}

\end{lemma}

Using Lemma~\ref{lem:convergencerates}, it can be seen that the epistemic uncertainty at the hyperparameter level, quantified by the sensitivity $I(U;\Zbf_{1:N})/Nm$, scales as $O(d \log(N) /N)$ for fixed $m$; while the epistemic uncertainty at the per-task level, accounted for by the sensitivity $ I(W;\Zbf|U)/m$, scales as $O(d \log (m)/m)$. Therefore,
 if $N \rightarrow \infty$, and $m$ is finite, the MEMR depends solely on the per-task epistemic uncertainty term $ I(W;\Zbf|U)/m$ in \eqref{eq:MER_upperbound}. That the MEMR does not vanish as $N \rightarrow \infty$ is a consequence of the fact that the meta-test task is a priori unknown. As a result, even an infinite amount of meta-training data does not resolve the epistemic uncertainty about the meta-test task \cite{gordon2018decision}, \cite{jose2021tasksimilarity}.

\subsection{Note on the Optimality of Bi-Level Meta-Learning}\label{sec:bilevel}

The meta-decision rule maps directly the observed meta-training set $\mset$, the training data $\Zbf$ of the meta-test task, and test feature input $X$ into a predictive distribution $q(y|X,\Zbf,\mset)$ on the space $\Yscr$ of target labels. By standard results in Bayesian inference (see \eg, \cite{bishop2006pattern}), the optimal predictive distribution is hence given by the posterior $P_{Y|X,\Zbf,\mset}$. To conclude this section and prepare for the next, we recall here that the joint distribution \eqref{eq:metalearning_distribution} can be factorized as
\begin{align}
    P_{Y|X,\Zbf,\mset}= \Ebb_{P_{U|\Zbf,\mset}P_{W|\Zbf,X,U}}[P_{Y|X,W}].\label{eq:ensemble_predictor}
\end{align} This factorization reveals that the optimal meta-decision rule can be implemented as a \textbf{two-step procedure}, whereby one first obtains the \textbf{hyperposterior} distribution  $ P_{U|\Zbf,\mset}$ using meta-training data $\mset$ and the meta-test task training data $\Zbf$; and then evaluates the \textbf{per-task posterior} distribution  $ P_{W|\Zbf,X,U}$ to evaluate the ensemble predictor \eqref{eq:ensemble_predictor}.
\subsection{Impact of Model Misspecification}
The results discussed so far rely on the assumption that the model is well specified, in the sense that the unknown population distribution belongs to a model class $\Mscr=\{P_{Z|w}:w \in \Wscr\}$. This assumption is violated if the true data generating distribution does not belong to the model class. In this subsection, we extend our results to this scenario.

To account for model misspecification, we assume that a task environment distribution $Q_T$ defines a distribution over tasks; and that each task $T_i \sim Q_T$, for $i=1,\hdots,N$, corresponds to a data distribution $Q_{Z|T_i}$, with its training data generated as $\Zbf_i \sim Q_{Z|T_i}^{\otimes m}$. The meta-training tasks $T_{1:N}$, the meta-training set $\mset$, the meta-test task $T$, the test task training data $\Zbf$, and the test data $Z$ are jointly distributed as
\begin{align}
    &Q_{T_{1:N},\mset,T,\Zbf,Z}\non \\&=\biggl(Q_T\otimes Q_{Z|T}^{\otimes m}\biggr)^{\otimes N} \otimes Q_T \otimes Q_{Z|T}^{\otimes m} \otimes Q_{Z|T}. \label{eq:joint2}
\end{align} Consequently, the observed meta-training data, meta-test training and test data are generated according to the marginal distribution $Q_{Y,X,\Zbf,\mset}$. Crucially, a learner that assumes this distribution computes the optimal ensemble predictor that minimizes the MEMR as $Q_{Y|X,\Zbf,\mset}$, whereas a learner assuming the marginal $P_{Y,X,\Zbf,\mset}$ under the joint distribution \eqref{eq:jointdistribution} computes it as $P_{Y|X,\Zbf,\mset}$. The model is misspecified in the sense that the marginal distributions $Q_{Y,X,\Zbf,\mset}$ and $P_{Y,X,\Zbf,\mset}$ differ.

We now analyze the impact of model misspecification on the Bayesian meta-risk \eqref{eq:Bayesianrisk_meta} under the log-loss. 
For a well-specified (WS) model, the
optimal ensemble predictor $P_{Y|X,\Zbf,\mset}$ (as in \eqref{eq:ensemble_predictor}) results in the following meta-risk 
\begin{align}
    R_{\log}^{WS}(Y|X,\mset,\Zbf)=\Ebb_{P_{\mset,\Zbf,X,Y}}[-\log P_{Y|X,\Zbf,\mset}].
\end{align}
When the model is misspecified (MS), the  ensemble predictor  $P_{Y|X,\Zbf,\mset}$ yields the following risk
\begin{align}
    R_{\log}^{MS}(Y|X,\mset,\Zbf)=\Ebb_{Q_{\mset,\Zbf,X,Y}}[-\log P_{Y|X,\Zbf,\mset}].
\end{align} The average excess risk due to model misspecification can be then quantified as
\begin{align}
    \Delta(Q,P)=R_{\log}^{MS}(Y|X,\mset,\Zbf)-R_{\log}^{WS}(Y|X,\mset,\Zbf).
\end{align} The overall minimum excess meta-risk of the ensemble predictor $P_{Y|X,\Zbf,\mset}$ under model misspecification is then given as
\begin{align}
    \MEMR_{\log}^{MS}=R_{\log}^{MS}(Y|X,\mset,\Zbf)-H(Y|X,W). \end{align}
    This can be decomposed as the sum of uncertainty due to model-misspecification and the epistemic uncertainty in prediction, \ie,
    \begin{align}
 \MEMR_{\log}^{MS}&=\Delta(Q,P)+\MEMR_{\log},
\end{align} where the second term, $\MEMR_{\log}$, was studied in the previous sections.
 
\section{Examples}\label{sec:examples}
\begin{figure}[t]
 \begin{center}
   \includegraphics[width=1\columnwidth]{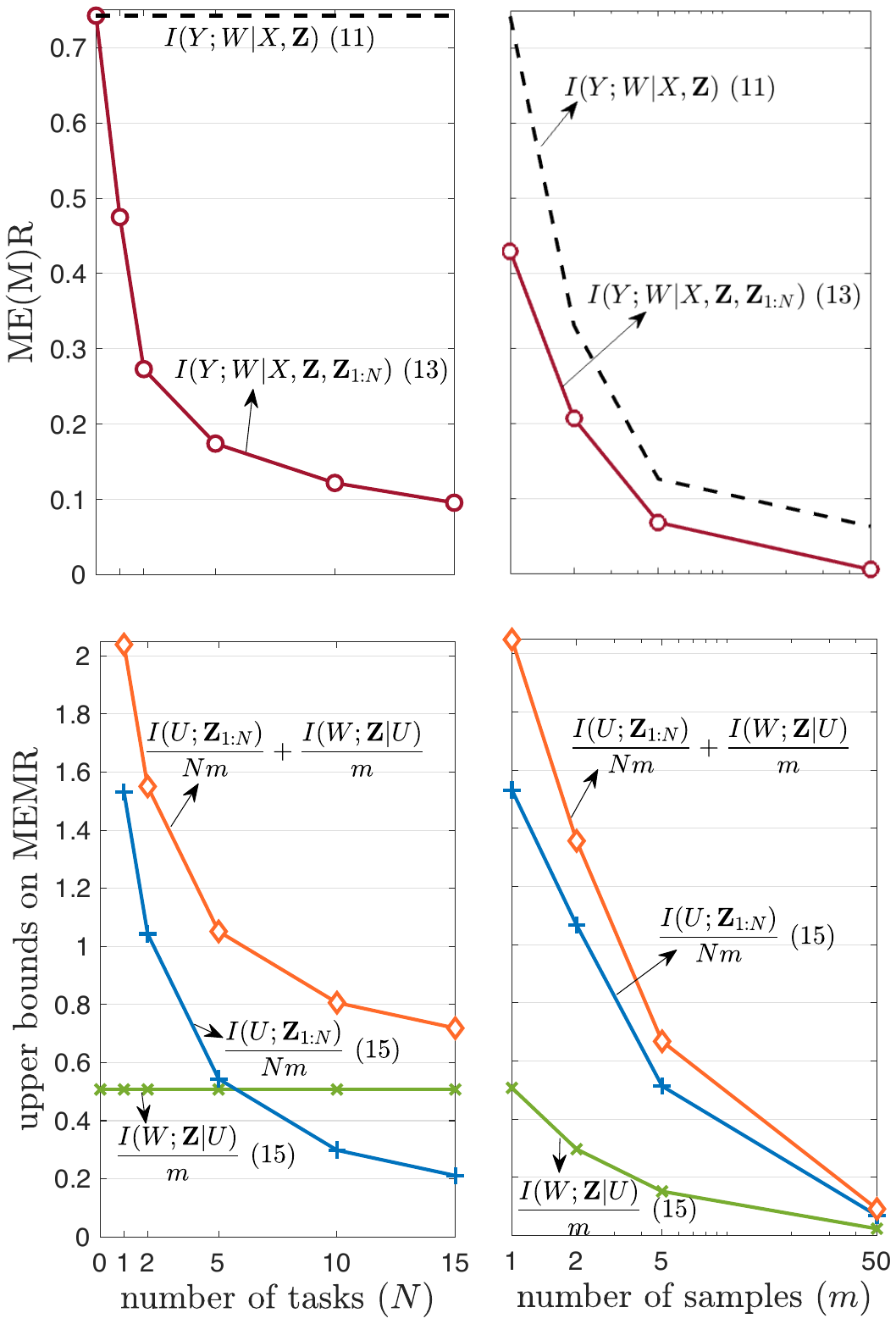} 
   \caption{MEMR (top) and information-theoretic upper bounds (bottom); (left) as a function of number of meta-training tasks ($N$) with fixed number of samples $m=1$ (right) as a function of number of per-task samples ($m$) with fixed number of meta-training tasks $N=1$. Experiments are evaluated with prior variance  $\sigma_w^2 = 0.01$ on the model parameters.} \label{fig:exp_small_std}
   \end{center}
\end{figure}  

In this section, we describe a regression example based on Bayesian neural networks \cite{neal2012bayesian}. Additional experiments can be found in supplementary material. 

In order to ensure that the model is well specified -- an underlying assumption of the analysis of generalization for Bayesian learning as studied in this work -- we consider a synthetic data set generated as follows.  We focus on a regression problem in which the target variable is distributed as $Y = f_W(X) + \xi$, with input $X \sim \mathcal{N}(0, 1)$; regression function $f_W(\cdot)$ specified by a neural network parameter vector $W $; and observation noise $\xi \sim \mathcal{N}(0, 0.1^2)$. The neural network consists of one hidden layer with ReLU activation in the hidden layer and a linear activation in the last layer. The prior distribution of model parameter $W$ is determined by hyperparameter $U$ as $P_{W|U}=\mathcal{N}(U, \sigma_w^2 \mathds{1})$ with fixed standard deviation $\sigma_w$ and identity matrix $\mathds{1}$ with the same dimension as vectors $W$ and $U$. Lastly, the hyperprior distribution for hyperparameter $U$ is defined as $P_U = \mathcal{N}(U|\mathbf{0}, \mathds{1})$ with an all-zero mean vector $\mathbf{0}$.

Fig.~\ref{fig:exp_small_std} compares the MEMR under the log-loss in \eqref{eq:MEMR_log} (top) and the information-theoretic upper bound $\MEMR^{\UB}_{\log}$ in \eqref{eq:MER_upperbound} (bottom) as a function of the increasing number $N$ of tasks for fixed $m=1$ (left) and of the increasing number of per-task samples $m$ for fixed $N=1$ (right). Note that when $N=0$, MEMR corresponds to the minimum excess risk for conventional learning. We use conditional MINE (C-MINE) \cite{mukherjee2020ccmi} along with Smoothed Mutual Information Lower-bound Estimator (SMILE) \cite{song2019understanding} to estimate the mutual information terms. Details for the experiment can be found in supplementary material.

The top part of the figure demonstrates the advantages of meta-learning over conventional learning as the number $N$ of meta-training tasks increases, and it also highlights the different dependence of the MEMR on $N$ and $m$. In particular, the MEMR does not vanish as $N$ grows larger, whereas having more per-task data, i.e., increasing $m$, yields a vanishing MEMR. This is due to the fact that,  even when one has access to infinitely many meta-training tasks, there is generally still some amount of unresolved uncertainty about the new meta-test task $\mathbf{Z}$  (see Lemma \ref{lem:convergencerates}). 

The bottom panels show the the information-theoretic upper bound $\MEMR^{\UB}_{\log}$ in \eqref{eq:MER_upperbound}, which separates the contributions to the MEMR due to epistemic uncertainty at the levels of hyperparameters and per-task model parameters. The bound, while numerically loose (see, e.g., \cite{hellstrom2020fast} and \cite{wang2021learning} for similar results), reproduce well the dependence of the MEMR on $N$ and $m$. Furthermore, the decomposition into the separate hyparparameter-level and model parameter-level contributions to epistemic uncertainty helps explain the non-vanishing behavior of the MEMR when $N$, as opposed to $m$, increases. While the hyperparameter-level sensitivity term decreases and vanishes asymptotically with $N$, the model parameter sensitivity term, which captures the uncertainty of the model parameter, is not influenced by $N$ and remains constant as $N$ varies. The non-vanishing MEMR can be thus attributed to the residual epistemic uncertainty about the newly encountered meta-test task at the level of model parameters.

The analysis of the two contributions is also useful to assess the relative merits of increasing $m$ or $N$. The left-bottom panel of Fig.~\ref{fig:exp_small_std} shows, for instance, that as $N$ increases, the contribution due to hyperparameter-level uncertainty  becomes less relevant than that of model parameter-level uncertainty. In this regime, further increases in $N$ have limited impact, and is generally preferable to increase $m$ (not shown).



\begin{figure}[t]
 \begin{center}
   \includegraphics[width=1\columnwidth]{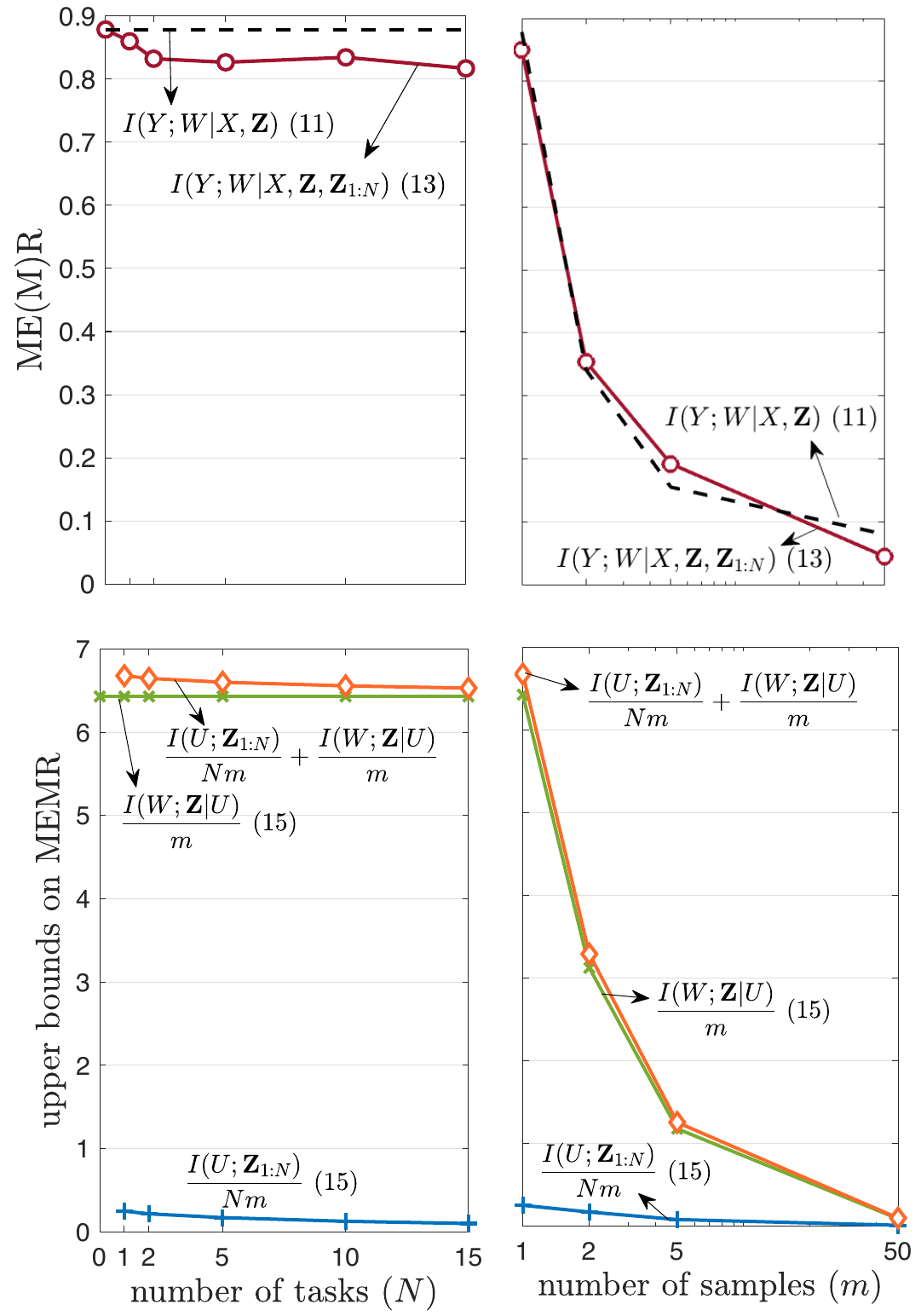} 
   \caption{MEMR (top) and information-theoretic upper bounds (bottom); (left) as a function of number of meta-training tasks ($N$) with fixed number of samples $m=1$ (right) as a function of number of per-task samples ($m$) with fixed number of meta-training tasks $N=1$. Experiments are evaluated with prior variance  $\sigma_w^2 = 1$ on the model parameters.} \label{fig:exp_large_std}
   \vspace{-0.7cm}
   \end{center}
\end{figure}  

In Figure~\ref{fig:exp_large_std}, we increase the prior variance $\sigma_w^2$ for the model parameters from $\sigma_w^2=0.01$, assumed in the previous figure, to $\sigma_w^2=1$. Intuitively, this change affects the amount of information that can be extracted from the hyperparameters $U$ on the model parameters and hence on the target variables. Accordingly, the model parameter sensitivity is seen to dominate the hyperparameter sensitivity, and meta-learning is observed to yield marginal benefits over conventional learning (\ie when $N=0$).

\section{Conclusion}
This paper studies epistemic uncertainty for Bayesian meta-learning from an information-theoretic perspective. We show that this uncertainty can be evaluated exactly (for log-loss) or bounded (for general loss functions) using a conditional MI $I(Y;W|X,\Zbf,\mset)$ involving model parameter, hyperparameter, and data. A novel information-theoretic upper bound on this term is also presented that explicitly shows the dependence of epistemic uncertainty on the number of meta-training tasks, $N$, and per-task samples, $m$.

The information-theoretic analysis conducted in this work assume  optimal Bayesian inference.  Future work may try to alleviate this limitations by considering the impact of approximations due to  variational inference. As a final note, as this paper addresses purely theoretical analysis, the results presented have no significant societal impact.

\def\aak#1{{{\color{red}#1}}}
\def\SC{{\rm SC}}
\def\OPT{{\rm OPT}}
\def\CC{{\rm CC}}
\def\UB{{\rm UB}}
\def\SimDP{{\rm SimDP}}
\def\LP{{\rm LP}}
\def\MH{{\rm MH}}
\def\DPI{{\rm DPI}}
\def\DP{{\rm DP}}
\def\sd{{\rm sd}}
\def\Lm{{\rm L}}
\def\tr{{\rm tr}}
\def\te{{\rm te}}
\def\Lh{{\widehat{ L}}}
\def\Lhm{{\widehat{ {\rm L}}}}
\def\Lsep{{{\rm L}^{\rm sep}_{\Zm_{1:N}}}}
\def\Ljnt{{{\rm L}^{\rm joint}_{\Zm_{1:N}}}}
\def\Ber{{\rm Bern}}
\def \Beta{{\rm Beta}}
\def \Bin{{\rm Binomial}}
\def\bbf{{\bf b}}
\def\KL{{\rm KL}}
\def\JS{{\rm JS}}
\def\base{{\rm base}}
\def\meta{{\rm meta}}
\def\Zm{S} 
\def \Pzt{P_{S|\tau}} 
\def \PzTt{P_{S|T=\tau}} 
\def \PzTtt{P_{S|T=T'}} 
\def \Pzti{P_{S|\tau_i}} 
\def \PzT{P_{S|T}} 
\def \PzTi{P_{S|T_i}} 
\def \PZmTi{P_{\Zm_i|\tau}} 
\def \PZMTi{P_{\Zm_i|T}} 
\def \PZmti{P_{\Zm_i|\tau_i}} 
\def \PZMti{P_{\Zm_i|T_i}} 
\def\DPSIED{{\rm DP-SIED}}
\def\DPSI{{\rm DPSI}}
\def\Xbf{\mathbf{X}}
\def\Ybf{\mathbf{Y}}
\def\mset{{\Zbf_{1:N}}}
\def\DPJE{{\rm DPJE}}
\def \Zbf{\mathbf{Z}}
\def \Ztbf{\mathbf{\widetilde{Z}}}
\def\JE{{\rm JE}}
\def\EWRM{{\rm EWRM}}
\def \excess {{\rm excess-risk}}
\def\trte{{\rm sep}}
\def\MER{{\rm MER}}
\def \MEMR{{\rm MEMR}}
\def\LPSI{{\rm LPSI}}
\def \explt{\exp(-\lambda(t|\Hscr(t))\Delta)}
\def\DPSWSID{{\rm DPSW-SID}}
\def\LPSWSID{{\rm LPSW-SID}}
\def\DPSWSIED{{\rm DPSW-SIED}}
\def\DPSWCSI{{\rm DPSW-CSI}}
\def\DPSWJE{{\rm DPSW-JE}}
\def\SW{{\rm SW}}
\def\ind{{\rm ind}}
\def\avg{{\rm avg}}
\def\Ztrain{Z^{\rm m_{tr}}}
\def\Ztest{Z^{\rm m_{te}}}
\def\Zjoint{Z^{\rm m_{jt}}}
\def\Ztr{D^{\rm m}}
\def \Zh{\widehat{Z}}
\def\mtr{m_{\rm tr}}
\def\mjt{m_{\rm jt}}
\def\Zte{D^{\rm m_{te}}}
\def\mte{m_{\rm te}}
\def\Dscrb{\bar{\mathcal{D}}}
\def\Dscrs{\mathcal{D}^{*}}
\def\Dscrt{\widetilde{\mathcal{D}}}
\def \Dscrtr{\mathcal{D}_{\rm{tr}}}
\def \Dscrte{\mathcal{D}_{\rm{te}}}
\def\Sscrm{\pmb{\mathscr{S}}}
\def\independenT#1#2{\mathrel{\rlap{$#1#2$}\mkern2mu{#1#2}}}
\def\csf{\mathsf{c}}
\def\ssf{\mathsf{s}}
\def\xm{\mathrm{x}}
\def \Im{\mathrm{I}}
\def \Tbf{\mathbf{D}}
\def\Ibf{\mathbf{I}}
\def\tfr{\rm{tfr}}
\def\Rbf{\mathbf{R}}
\def\lbf{\mathbf{l}}
\def\Abf{\mathbf{A}}
\def\Sbf{\mathbf{S}}
\def\Zbf{\mathbf{Z}}
\def\Zbft{\mathbf{\widetilde{Z}}}
\def\Iscrm{\mathcal{I}}
\def\Itt{\mathtt{I}}
\def \Sm{\mathrm{S}}
\def\Escr{\mathscr{E}}
\def\lt{\widetilde{l}}
\def\Lh{\widetilde{L}}
\def\thetat{\widetilde{\theta}}
\def\Dscrt{\widetilde{\mathcal{D}}}
\def\lambdat{\widetilde{\lambda}}
\def\Qt{\widetilde{Q}}
\def\Yt{\widetilde{Y}}
\def\qt{\widetilde{q}}
\def\kt{\widetilde{k}}
\def\pt{\widetilde{p}}
\def\zt{\widetilde{z}}
\def\phit{\widetilde{\phi}}
\def\mubf{\mathbf{\mu}}
\def \Nscr{\mathcal{N}}
\def \Wscr{\mathscr{W}}
\def \ge{\mathrm{gen-error}}
\def \DL{ {\rm \Delta} L}
\def \DLm{ {\rm \Delta} {\rm L}}
\def \lh{\widehat{l}}
\def \Lh{\widehat{L}}
\def \mge{\mathrm{metagen-error}}
%
%




%

%

\onecolumn
\appendix
\section{Proof of Lemma~4.1}\label{app:properties_MEMR}
The properties are a direct consequence of the data processing inequality satisfied by the Bayesian risk \cite{xu2020minimum}. This states that, given jointly distributed random variables $A,B$ and $C$, if the Markov chain $A-B-C$ holds, we have the inequality $R_{\ell}(C|A) \geq R_{\ell}(C|B)$. Noting that $(\mset,\Zbf,X) - (U,W,X)-Y$ forms a Markov chain, the non-negativity of the MEMR follows from the data processing inequality. Through the same argument, it can be proved that MEMR is non-increasing with the number of tasks and per-task data samples.
\section{Proof of Proposition~4.1}\label{app:MEMR_log}
 Under log-loss, the Bayesian meta-risk is given by,
\begin{align}
    R_{\log}(Y|X,\Zbf,\mset)=\min_{q(\cdot)}\Ebb_{P_{\mset,\Zbf,X}P_{Y|X,\Zbf,\mset}}[-\log q(Y|X,\Zbf,\mset)].
\end{align} From standard results in information theory \cite{cover06elements}, it can be verified that the optimal meta-decision rule $q(\cdot)$ that minimizes the Bayesian meta-risk corresponds to the posterior predictive distribution $P_{Y|X,\Zbf,\mset}$, whereby we have
\begin{align}
    R_{\log}(Y|X,\Zbf,\mset)=\Ebb_{P_{\mset,\Zbf,X}P_{Y|X,\Zbf,\mset}}[-\log P_{Y|X,\Zbf,\mset}] =H(Y|X,\Zbf,\mset).
\end{align} Similarly, it can be shown that
\begin{align}
    R_{\log}(Y|X,W,U)&=\Ebb_{P_{U,W,X}P_{Y|X,U,W}}[-\log P_{Y|X,U,W}] \non \\&=H(Y|X,W,U)=H(Y|X,W),
\end{align}where the last equality follows since $U-W-Z$ forms a Markov chain whereby $P_{Z|W,U}=P_{Z|W}$. Together, we then have that
\begin{align}
    \MEMR_{\log}&=H(Y|X,\Zbf,\mset)-H(Y|X,W)\non \\
    &\stackrel{(a)}{=}H(Y|X,\Zbf,\mset)-H(Y|X,W,\Zbf,\mset) \\
    &=I(Y;W|X,\Zbf,\mset),
\end{align}where the equality in $(a)$ follows since conditioned on test input $X$ and model parameter $W$, the test output $Y$ is independent of $(\Zbf,\mset)$.
\section{Proof of Theorem~4.2}\label{app:loglossUB}
To obtain the required bound on MEMR, we note that the following set of relations hold.
\begin{align}
\MEMR_{\log}&=I(W;Y|X,\Zbf,\Zbf_{1:N}) \non \\
&\leq I(W;Z|\Zbf,\Zbf_{1:N}) \label{eq:11} \\
&\leq \frac{I(W;\Zbf|\Zbf_{1:N})}{m}
\label{eq:33}\\
&=\frac{I(W,U;\Zbf|\Zbf_{1:N})}{m} \label{eq:44}\\
&=\frac{I(W;\Zbf|U)}{m}+ \frac{I(U;\Zbf|\Zbf_{1:N})}{m} \\
&\leq \frac{I(W;\Zbf|U)}{m}+ \frac{I(U;\Zbf_{1:N})}{Nm} \label{eq:55}.
\end{align}
Here, \eqref{eq:11} follows from the chain rule of mutual information. 
To prove inequality \eqref{eq:33}, we use the technique of
\cite[Proof of Thm.~2]{xu2020minimum} which we now detail here. Towards this, we define $Z^j=(Z_1,\hdots,Z_j)$ and
note that $I(W;\Zbf|\Zbf_{1:N})=\sum_{j=1}^m I(W;Z_j|Z^{j-1},\Zbf_{1:N})$. We then have the following set of relations
\begin{align}
&I(W;Z_{j+1}|Z^j,\Zbf_{1:N})\non \\&=H(Z_{j+1}|Z^j,\Zbf_{1:N})-H(Z_{j+1}|Z^j,W,\Zbf_{1:N})\non \\
&\stackrel{(a)}{=}H(Z_{j+2}|Z^j,\Zbf_{1:N})-H(Z_{j+1}|W) \non \\
& \stackrel{(b)}{\geq} H(Z_{j+2}|Z^{j+1},\Zbf_{1:N})-H(Z_{j+2}|W) \non \\
&=H(Z_{j+2}|Z^{j+1},\Zbf_{1:N})-H(Z_{j+2}|\Zbf_{1:N},W,Z^{j+1}) \non \\
&=I(W;Z_{j+2}|\Zbf_{1:N},Z^{j+1}),
\end{align}
where $(a)$ follows since $(Z_{j+1},Z^j,\Zbf_{1:N}) \triangleq (Z_{j+2},Z^j,\Zbf_{1:N})$ in distribution and that $Z_{j+1}$ is conditionally independent of $(Z^j,\mset)$ given $W$, and $(b)$ follows since conditioning reduces entropy, and that $(Z_{j+1},W) \triangleq (Z_{j+2},W)$ in distribution. Consequently, we get the inequality
\begin{align}
I(W;\Zbf|\Zbf_{1:N})&=\sum_{j=1}^m I(W;Z_j|Z^{j-1},\Zbf_{1:N})\non \\& \geq m I(W;Z_{m+1}|Z^{m},\Zbf_{1:N})\non \\&=m I(W;Z|\Zbf,\Zbf_{1:N}),
\end{align}whereby $I(W;Z|\Zbf,\Zbf_{1:N}) \leq I(W;\Zbf|\Zbf_{1:N})/m$.

The equality in \eqref{eq:44} follows since $I(U;\Zbf|W,\mset)=0$, which results from $(U,\mset)-W-\Zbf$ forming a Markov chain, whereby $I(W,U;\Zbf|\mset)=I(W;\Zbf|\mset)$.
Finally to see \eqref{eq:55}, we follow similar steps as in the proof of \eqref{eq:33}. Denoting $\Zbf^{k}=(\Zbf_1,\hdots,\Zbf_k)$, we have the mutual information $I(U;\mset)=\sum_{k=1}^N I(U;\Zbf_k|\Zbf^{k-1})$, each individual component of which can be written as
\begin{align}
&I(U;\Zbf_{k+1}|\Zbf^k)\non \\
&=H(\Zbf_{k+1}|\Zbf^k)-H(\Zbf_{k+1}|U,\Zbf^k)\non \\
&\stackrel{(a)}{=}H(\Zbf_{k+2}|\Zbf^k)-H(\Zbf_{k+1}|U) \\
&\stackrel{(b)}{\geq} H(\Zbf_{k+2}|\Zbf^{k+1})-H(\Zbf_{k+2}|U) \non \\
&=H(\Zbf_{k+2}|\Zbf^{k+1})-H(\Zbf_{k+2}|U,\Zbf^{k+1})\non\\
&=I(\Zbf_{k+2};U|\Zbf^{k+1}).
\end{align} Here, the equality in $(a)$ follows since $(\Zbf_{k+1},\Zbf^k)\triangleq (\Zbf_{k+2},\Zbf^k)$ in distribution, and that $\Zbf_{k+1}$ is conditionally independent of $\Zbf^k$ given $U$, and $(b)$ follows since conditioning reduces entropy, and that $(\Zbf_{k+1},U) \triangleq (\Zbf_{k+2},U)$ in distribution. Consequently, we have that the mutual information $I(U;\mset)\geq N I(U;\Zbf|\mset)$, which results in the inequality in \eqref{eq:55}.
\section{Meta-Learning vs Conventional Learning}\label{app:meta-conventional}
One of the advantages of the Bayesian viewpoint on meta-learning is that one can obtain general information-theoretic conclusions about the performance comparison of meta-learning and conventional learning. This is in contrast to the frequentist analyses that focus on the meta-generalization error \cite{baxter2000model}, \cite{maurer2005algorithmic},  \cite{pentina2014pac}, making it difficult to draw general conclusions on this comparison.

To start, it is easy to see that under the assumption $\mset=\emptyset$ and $P_{W|U}=P_W$, the MEMR (defined in (10) in the main text) reduces to the MER (equation (5) in the main text) for conventional Bayesian learning. In fact,  we have 
%
$
\MEMR_{\log}=I(W;Y|X,\Zbf)
=\MER_{\log}.
$

Generalizing this observation, the following proposition quantifies the gains of meta-learning with respect to conventional learning under log-loss. 
\begin{proposition}\label{prop:1}
Under log-loss, the $\MEMR_{\log}$  for meta-learning and the $\MER_{\log}$ for conventional learning
are related as
\begin{align}
    \MER_{\log}-\MEMR_{\log}&=I(\mset;Y|X,\Zbf) \geq 0. \label{eq:gain1}
\end{align}
\end{proposition}
\begin{proof}
The relation in \eqref{eq:gain1} is obtained as follows.
\begin{align}
    \MER_{\log}-\MEMR_{\log}&=I(Y;W|X,\Zbf)-I(Y;W|X,\Zbf,\mset)\\
    &=H(Y|X,\Zbf)-H(Y|X,\Zbf,\mset)\\
    &=I(Y;\mset|X,\Zbf).
    \end{align}
\end{proof}

Proposition~\ref{prop:1} shows that under the log-loss, meta-learning yields a lower minimum excess risk than conventional learning. The gain in minimum excess risk is quantified by the conditional MI $I(\mset;Y|X,\Zbf)$, which grows as the meta-training set $\mset$ becomes more informative about the meta-test target variable $Y$ beyond the information already available in the meta-test training set $\Zbf$ and input $X$.
\section{Assumptions for the Convergence Rates of Lemma~4.3}\label{app:convergence_rates}
In this section, we specialize the assumptions required for the convergence rate of $I(W;\Zbf)$ for Bayesian learning in \cite{clarke1994jeffreys} to the case of Bayesian meta-learning, where we have two MI terms $I(U;\Zbf_{1:N})$ and $I(W;\Zbf|U)$.
We first list the assumptions required for the convergence of $I(U;\Zbf_{1:N})$, and explain how these extend to $I(W;\Zbf|U)$.
\begin{assumption}\label{assum:convergencerates}
The following assumptions must be satisfied for ensuring the convergence of the mutual information term $I(U;\mset)$ in Lemma~4.3.
\begin{enumerate}
    \item Let $U \in \Uscr \subset \Real^d$, and that the density $P_{\Zbf|U}$ exists with respect to Lebesgue measure. Moreover, $\Uscr$ has a non-void interior and its boundary has $d$-dimensional Lebesgue measure 0.
    \item The density $P_{\Zbf|U}(\Zbf|u)$ is twice continuously differentiable in $u$ for almost every $\Zbf$ and there exists $\delta(u)$ so that for each $j,k =1,\hdots,d$,
    \begin{align}
       f(u)= \Ebb_{P_{\Zbf|U=u}}\biggl[\sup_{u':||u'-u||<\delta(u)} \biggl|\frac{\partial^2}{\partial u_j' \partial u_k'}  \log P_{\Zbf|U}(\Zbf|u')\biggr|^2 \biggr]
    \end{align}
    is finite and continuous.
    \item For $j=1,\hdots,d$,
    \begin{align}
        \Ebb_{P_{\Zbf|U=u}} \biggl[\biggl| \frac{\partial}{\partial u_j} \log P_{\Zbf|U}(\Zbf|u) \biggr|^{2+\xi} \biggr]
    \end{align} is finite and continuous, as a function of $u$, for some $\xi>0$.
    \item Fisher information matrix (FIM) and second derivative of relative entropy are equal \ie for matrices,
    \begin{align}
        [I_{\Zbf|U}(u)]_{j,k}=\Ebb \biggl[\frac{\partial}{\partial u_j} \log P_{\Zbf|U}(\Zbf|u)\frac{\partial}{\partial u_k} \log P_{\Zbf|U}(\Zbf|u)  \biggr]
    \end{align}
  and 
  \begin{align}
      [J_{\Zbf|U}(u)]_{j,k}=\biggl[ \frac{\partial^2}{\partial u_j' \partial u_k'} D_{\KL}(P_{\Zbf|u}||P_{\Zbf|u'}) \biggr|_{u'=u}\biggr],
      \end{align}
      we have $I_{\Zbf|U}(u)=J_{\Zbf|U}(u)$ and that the matrix $I_{\Zbf|U}(u)$ is assumed to be positive definite.
      \item For $u \neq u'$, we have $P_{\Zbf|U=u} \neq P_{\Zbf|U=u'}.$
      \item The hyperprior $P_U$ is assumed continuous and is supported on a compact subset in the interior of $\Uscr.$
\end{enumerate}
\end{assumption}
Under Assumption~\ref{assum:convergencerates}, Theorem~1 of \cite{clarke1994jeffreys} then yields the required asymptotic of the MI $I(U;\mset)$ in Lemma~4.3.

To analyze the asymptotic of the MI $I(W;\Zbf|U)$, we note that $I(W;\Zbf|U)=\Ebb_{P_U}[I(W;\Zbf|U=u)]$. Consequently, we specialize Assumption~\ref{assum:convergencerates} to ensure convergence of $I(W;\Zbf|U=u)$ for each $u \in \Uscr$. This can be done by replacing the  distribution $P_{\Zbf|U}$ with $P_{Z|W}$, the hyperprior $P_U$ by the prior $P_{W|U=u}$ for each $u \in \Uscr$, such that the resulting assumptions hold at the level of model parameter. Subsequently, Theorem~1 of \cite{clarke1994jeffreys} ensures that as $m \rightarrow \infty$,
\begin{align}
    I(W;\Zbf|U=u)=\frac{d}{2}\log \Bigl(\frac{m}{2\pi e}\Bigr)+H(W|U=u)+\Ebb_{ P_{W|U=u}} \Bigl[ \log |J_{Z|W}(W)| \Bigr]+o(1). \label{eq:1}
\end{align}Taking expectation of \eqref{eq:1} with respect to the hyperprior $P_U$, then yields the asymptotic behaviour of $I(W;\Zbf|U)$ in Lemma~4.3.

\section{Information-Theoretic Analysis of the MEMR for General Loss Functions}\label{sec:generalloss_app}
In this section, we extend the characterization in Theorem~4.2 of the MEMR from the log-loss to general loss functions $\ell:\Yscr \times \Ascr \rightarrow \Real$. We specifically show that, under suitable assumptions on the loss function, the MEMR (equation (10) in the main text) can be upper bounded using a concave, non-decreasing, function of the conditional mutual information $I(Y;W|X,\Zbf,\mset)$.

To upper bound the $\MEMR_l$, we consider the performance of the following randomized meta-decision rule  $\Psi_{\meta}(X,\Zbf,\mset)$. Define as $\Phi^{*}_{\meta}( X ,W,U)$ the optimal genie-aided decision rule that minimizes the Bayesian meta-risk, \ie,
 $
    R_{\ell}(Y|X,W,U)=\Ebb[\ell(Y,\Phi^{*}_{\meta}(X,W,U))].
$ This rule is not directly applicable since the pair $(U,W)$ is not known. Having computed the posterior $P_{W,U|X,\Zbf,\Zbf_{1:N}}$ (see Section~4.2 of the main text), we draw a sample $(U',W')$ from it.  Note that conditioned on $(X,\Zbf,\mset)$, the pairs $(U,W)$ and $(U',W')$ are independent. The meta-decision rule is chosen as $\Phi_{\meta}^{*}(X,W',U')$, substituting the true pair $(U,W)$ with the sample $(U',W')$.
Consequently, the MEMR can be upper bounded as
\begin{align}
\MEMR_l  \leq \Ebb_{P_{X,\Zbf,\Zbf_{1:N}}P_{Y,W',U'|X,\Zbf,\Zbf_{1:N}}}[\ell(Y,\Phi^{*}_{\meta}(X,W',U'))]-\Ebb_{P_{X,Y,W,U}}[\ell(Y,\Phi^{*}_{\meta}(X,W,U))]. \label{eq:ub_1}
\end{align}

We now obtain an information-theoretic upper bound on \eqref{eq:ub_1} under the following assumption. Towards this, we first define the following zero mean random variable
\begin{align}
&\Delta \ell(Y,W',U'|X,\Zbf,\Zbf_{1:N})\non =\ell(Y,\Phi^{*}_{\meta}(X,W',U')-\Ebb_{P_{Y,W',U'|X,\Zbf,\Zbf_{1:N}}}[\ell(Y,\Phi^{*}_{\meta}(X,W',U')]).
\end{align}
\begin{assumption}\label{assum:1}
There exists function $\Upsilon(\lambda)$ for $\lambda \in (0,b]$ satisfying $\Upsilon(0)=\Upsilon'(0)=0$ such that the cumulant generating function (CGF) of $\Delta \ell(Y,W',U'|x,\zbf,\zbf_{1:N})$ is upper bounded by $\Upsilon(\lambda)$, \ie, the following inequality holds
\begin{align}
\log \Ebb_{P_{Y,W',U'|x,\zbf,\zbf_{1:N}}}&\biggl[\exp \biggl(\lambda\Delta \ell(Y,W',U'|x,\zbf,\zbf_{1:N})\biggr)\biggr] \leq \Upsilon(\lambda)
\end{align} for all $x \in \Xscr,$  $\zbf \in \Zscr^m$ and $\zbf_{1:N}\in \Zscr^{Nm}$. 
\end{assumption}

We also define the Legendre dual of $\Upsilon(\lambda)$ as $\Upsilon^{*}(x)=\sup_{\lambda \in (0,b]}\lambda x -\Upsilon(\lambda)$. It is a non-negative, convex and a non-decreasing function on $[0,\infty)$ with $\Upsilon^{*}(0)=0$ \cite[Lemma~2.4]{boucheron2013concentration}. The inverse of this function, called inverse Legendre dual, is defined as  $\Upsilon^{*-1}(y)=\inf_{\lambda \in (0,b]}( y+ \Upsilon(\lambda))/\lambda$, and is concave. We now state our result.
\begin{theorem}\label{thm:generalloss}
Under Assumption~\ref{assum:1}, the following bound on MEMR holds
\begin{align}
\MEMR_{l} &\leq \Upsilon^{*-1}\biggl(I(W;Y|X,\Zbf,\Zbf_{1:N} \biggr)\label{eq:MEMR_loss} \\
& \leq \Upsilon^{*-1}\biggl(\frac{I(U;\Zbf_{1:N})}{Nm}+\frac{I(W;\Zbf|U)}{m}\biggr).
\end{align}
\end{theorem}
\begin{proof}
The proof follows the approach in \cite{xu2020minimum} and we outline the main steps here. The following set of relations hold:
\begin{align}
& \MEMR_l\non \\&\leq    \Ebb_{P_{X,\Zbf,\Zbf_{1:N}}P_{Y,W',U'|X,\Zbf,\Zbf_{1:N}}}[\ell(Y,\Phi^{*}_{\meta}(X,W',U'))]-\Ebb_{P_{X,Y,W,U}}[\ell(Y,\Phi^{*}_{\meta}(X,W,U))] \non \\
&=\Ebb_{P_{X,\Zbf,\Zbf_{1:N}}}\biggl[\Ebb_{P_{Y,W',U'|X,\Zbf,\Zbf_{1:N}}}[\ell(Y,\Phi^{*}_{\meta}(X,W',U'))]-\Ebb_{P_{Y,W,U|X,\Zbf,\mset}}[\ell(Y,\Phi^{*}_{\meta}(X,W,U))] \biggr]\non\\
&\stackrel{(a)}{\leq } \Ebb_{P_{X,\Zbf,\Zbf_{1:N}}}\biggl[\Upsilon^{*-1}\biggl(D_{\KL}(P_{Y,W,U|X,\Zbf,\mset}||P_{Y,W',U'|X,\Zbf,\mset}) \biggr)\biggr] \non \\
& \stackrel{(b)}{\leq}\Upsilon^{*-1}\biggl(\Ebb_{P_{X,\Zbf,\Zbf_{1:N}}}\Bigl[D_{\KL}(P_{Y,W,U|X,\Zbf,\mset}||P_{Y,W',U'|X,\Zbf,\mset})\Bigr] \biggr) \non \\
&\stackrel{(c)}{=}\Upsilon^{*-1}\biggl(I(Y;W,U|X,\Zbf,\mset) \biggr)\non \\
&=\Upsilon^{*-1}\biggl(I(Y;W|X,\Zbf,\mset) \biggr)\non \\
& \stackrel{(d)}{\leq} \Upsilon^{*-1}\biggl(\frac{I(U;\Zbf_{1:N})}{Nm}+\frac{I(W;\Zbf|U)}{m}\biggr).
\end{align}
Here, the inequality in $(a)$ follows from Assumption~\ref{assum:1} and using Donsker-Varadhan inequality (see \cite[Lemma~A.1]{xu2020minimum}). The inequality in $(b)$ follows by using Jensen's inequality on the concave inverse Legendre dual function $\Upsilon^{*-1}(\cdot)$. The equality in $(c)$ follows from the observation  that while the distribution  $P_{Y,W,U|X,\Zbf,\mset}$ factorizes as $P_{Y|X,\Zbf,\mset} \otimes P_{W,U|X,Y,\Zbf,\mset}$, the distribution $P_{Y,W',U'|X,\Zbf,\mset}$ is obtained as $P_{Y|X,\Zbf,\mset} \otimes P_{W,U|X,\Zbf,\mset}$ with $(W',U')$ conditionally independent of $Y$ given $(X,\Zbf,\mset)$. The last inequality in $(d)$ follows since the inverse Legendre dual is a non-decreasing function.
\end{proof}

It can be seen that if the random variable $\Delta \ell (Y,W',U'|x,\zbf,\zbf_{1:N})$ is $\sigma^2$-sub Gaussian\footnote{A zero-mean random variable $X$ is said to be $\sigma^2$-sub-Gaussian if the cumulant generating function (CGF) satisfies $\log \Ebb_{P_X}[\exp(\lambda(X))]\leq \frac{\lambda^2\sigma^2}{2}$ for all $\lambda \in \Real$.} when $(Y,W',U') \sim P_{Y,W',U'|x,\zbf,\zbf_{1:N}}$ for all $x \in \Xscr$, $\zbf \in \Zscr^m$ and $\zbf_{1:N} \in \Zscr^{Nm}$, then Assumption~\ref{assum:1} is satisfied with $b=\infty$, $\Upsilon(\lambda)=\lambda^2\sigma^2/2$ and $\Upsilon^{*-1}(y)=\sqrt{2\sigma^2 y}$. We now specialize Theorem~\ref{thm:generalloss} to account for this case.
\begin{corollary}
Assume that $\Delta \ell (Y,W',U'|x,\zbf,\zbf_{1:N})$ is $\sigma^2$-sub Gaussian for all $x \in \Xscr$, $\zbf \in \Zscr^m$ and $\zbf_{1:N} \in \Zscr^{Nm}$. Then, the following upper bound on the MEMR holds:
\begin{align}
    \MEMR_l
    &\leq \sqrt{2\sigma^2 \biggl(\frac{I(U;\Zbf_{1:N})}{Nm}+\frac{I(W;\Zbf|U)}{m}\biggr) }.
\end{align}
\end{corollary}
\section{Additional Experiments}
\subsection{Bayesian Sinusoidal Regression}
We focus on a sinusoidal regression problem, in which we have $Y=W \sin(X)+\xi$, with amplitude $W$ and observation noise $\xi \sim \Nscr(0, 1)$. The prior distribution of model parameter $W$ is determined via hyperparameter $U$ as $P_{W|U}=\Nscr(W|U,\sigma_w^2)$ for some fixed variance $\sigma^2_w$. The hyperprior on the prior-mean $U$ is taken as $P_U=\Nscr(0,1)$. 

\begin{figure}[h!]
 \begin{center}
   \includegraphics[width=1\columnwidth, trim=0.2in 1.3in 0in 1.3in,clip=true]{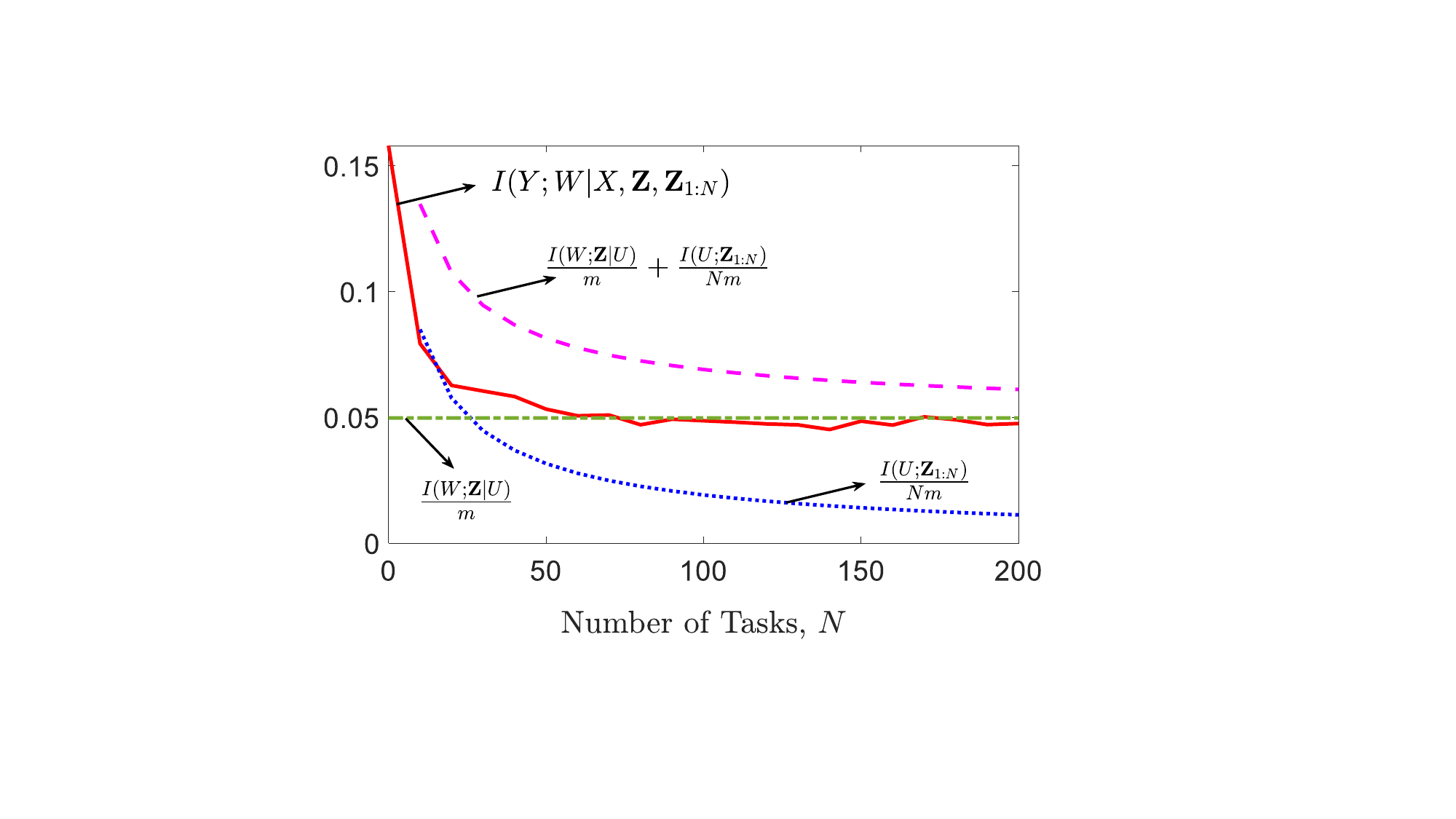} 
   \caption{MEMR (13), information-theoretic upper bound (15) and the sensitivity terms $I(W;\Zbf|U)/m$ and $I(U;\Zbf_{1:N})/Nm$ as a function of number of meta-training tasks ($N$) with fixed number of samples $m=1$ for $\sigma^2_w=0.2$.} \label{fig:sinusoidal_expt1_fig1}
   \end{center}
\end{figure}  
Figure~\ref{fig:sinusoidal_expt1_fig1} compares the MEMR under log loss in (13) with the upper bound $\MEMR_{\log}^{\UB}$ in (15), as well as its two component sensitivity terms -- model parameter-level sensitivity $I(W;\Zbf|U)/m$  and hyperparameter-level sensitivity $I(U;\Zbf_{1:N})/Nm$ -- as a function of the increasing number $N$ of tasks for fixed $m=1$ and $\sigma^2_w=0.2$. Note that when $N=0$, MEMR corresponds to the minimum excess risk for conventional learning. Figure~\ref{fig:sinusoidal_expt1_fig1} shows that for the problem setting studied, meta-learning using large number $N$ of tasks can yield significantly lower excess risk than conventional learning. In fact, while increasing $N$ decreases both MEMR and the corresponding upper bound at first, they remain non-vanishing in the limit of large number of tasks. This can be explained by looking at the two sensitivity terms - while the hyper-parameter level sensitivity term  decreases and vanishes asymptotically, the model parameter sensitivity term, which captures the uncertainty due to limited number $m$ of samples observed about the new, previously unobserved meta-test task, is not influenced by $N$ and remains constant. The non-vanishing MEMR can be thus attributed to the residual epistemic uncertainty about the newly encountered meta-test task.
\begin{figure}[h!]
 \begin{center}
   \includegraphics[width=1\columnwidth, trim=0.2in 1.3in 0in 1.3in,clip=true]{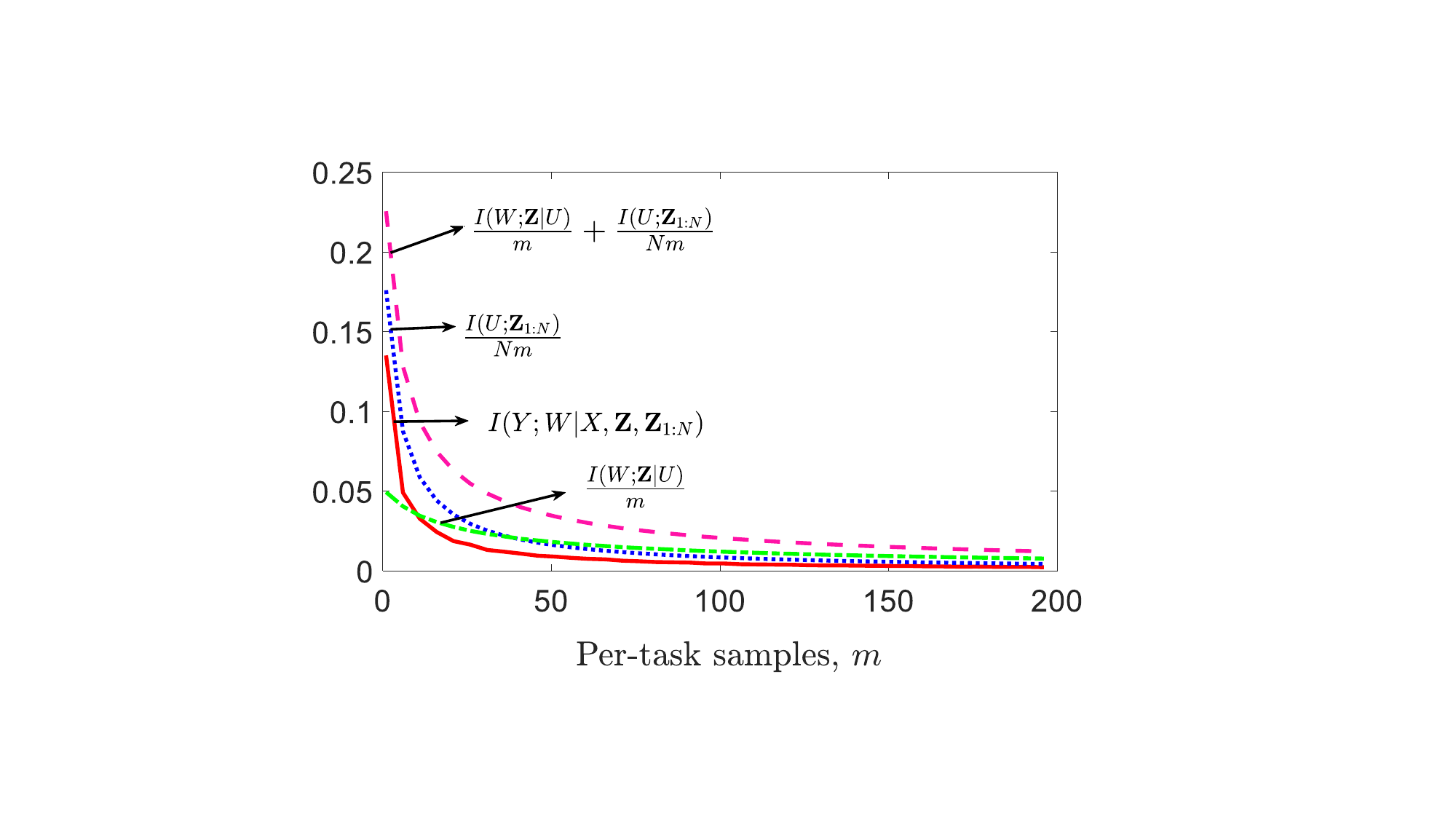} 
   \caption{MEMR (13), information-theoretic upper bound (15) and the sensitivity terms $I(W;\Zbf|U)/m$ and $I(U;\Zbf_{1:N})/Nm$ as a function of number per-task samples ($m$) with fixed number of meta-training tasks $N=1$ for $\sigma^2_w=0.2$.} \label{fig:sinusoidal_expt1_fig2}
   \end{center}
\end{figure} 

In Figure~\ref{fig:sinusoidal_expt1_fig2}, we compare the MEMR (13), the upper bound (15) and the two sensitivity terms as a function of increasing number $m$ of per-task samples, for fixed $N=1$ and $\sigma_w^2=0.2$. It can be seen that availability of abundant number of per-task training samples decreases the MEMR, the corresponding upper bound as well as the two sensitivity terms, all of which vanish asymptotically. When large number $m$  (say $m>75$) of training samples is available, the MEMR is largely determined by the model parameters-level sensitivity. As observed in Figure~\ref{fig:sinusoidal_expt1_fig1}, meta-learning using large $N$ has  no impact on model parameter sensitivity, and is thus not very beneficial over conventional learning in this regime. However, when $m$ is small (say $m<10$), it can be seen that the major contributor to the MEMR is the hyperparameter-level sensitivity term. Consequently, in this regime, meta-learning using large number $N$ of tasks can significantly reduce the hyperparameter sensitivity and thus the MEMR. 
\begin{figure}[h!]
 \begin{center}
   \includegraphics[width=1\columnwidth, trim=0.2in 1.8in 0in 1.3in,clip=true]{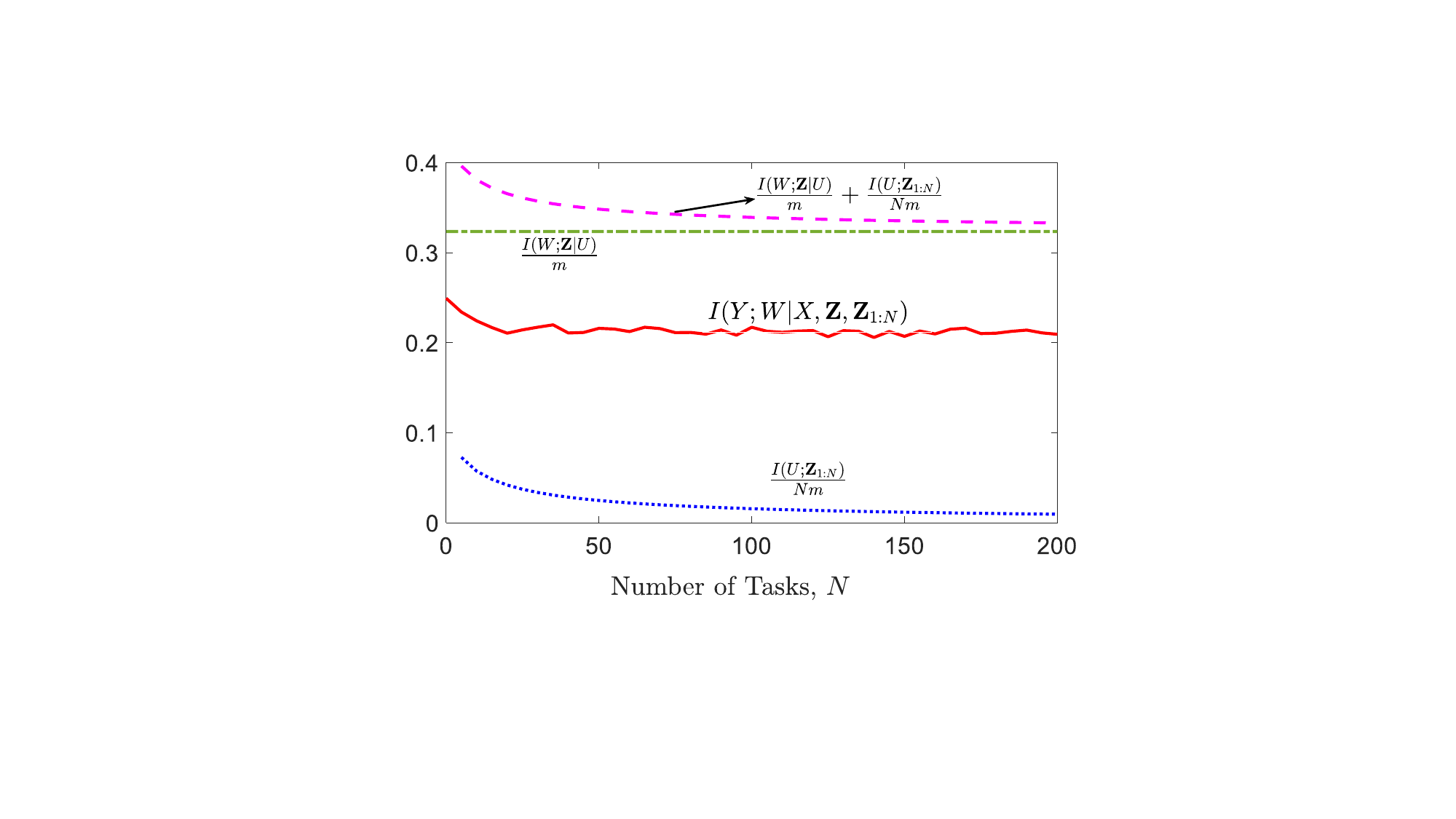} 
   \caption{MEMR (13), information-theoretic upper bound (15) and the sensitivity terms $I(W;\Zbf|U)/m$ and $I(U;\Zbf_{1:N})/Nm$ as a function of number per-task samples ($m$) with fixed number of meta-training tasks $N=1$ for $\sigma^2_w=2$.} \label{fig:sinusoidal_expt2_fig1}
   \end{center}
\end{figure} 

In Figure~\ref{fig:sinusoidal_expt2_fig1}, we increase the prior variance to $\sigma^2_w=2$ and compare the MEMR, the upper bound and the two sensitivity terms, as a function of the increasing number of meta-training tasks $N$ for fixed $m=1$. Recall that in Figure~\ref{fig:sinusoidal_expt1_fig1}, we considered the prior $P_{W|U}$ to be more concentrated at the mean or hyperparameter $U$ (by choosing small $\sigma_w^2=0.2)$. This results in lower model parameter sensitivity (given knowledge of $U$), while the hyperparameter sensitivity is large and is shown to decrease when meta-learning using large $N$. In contrast, by assuming a larger prior variance, the model parameter sensitivity outweighs the hyperparameter sensitivity. As such, meta-learning  using  large number $N$ of tasks brings marginal benefits over conventional learning (\ie when $N=0$) as shown by the  MEMR curve.

\section{Experimental Details for Bayesian Neural Network Regression}
\label{app:sin_regression_exp}
We detail the essential experimental settings for reproducibility of the results.\\
\begin{table}[ht!]
\centering
\begin{tabular}{c c} 

  \textbf{Settings} & \textbf{Value} \\ [0.5ex] 
 \hline
 regression architecture & MLP \\ 
 \# hidden layers & 1\\ 
 \# hidden units & 3 \\
 activation & ReLU \\[1ex] 
\end{tabular}
\caption{BNN architecture for Bayesian neural network regression}
\label{table_exp:1_1}
\end{table}

\begin{table}[ht!]
\centering
\begin{tabular}{c c} 

  \textbf{Settings} & \textbf{Value} \\ [0.5ex] 
 \hline
 classifier architecture for C-MINE & MLP + Sigmoid layer \\ 
 \# hidden layers & 3\\ 
 \# hidden units & (64,64,1) \\
 step size & 0.001  \\
 optimizer & Adam ($\beta_1=0.9, \beta_2=0.999$)  \\
 \# epoch & 200 \\
 batch size & 64 \\
 regularizer & L2 (0.001) \\
 activation & ReLU \\[1ex] 
\end{tabular}
\caption{classifier details in C-MINE for Bayesian neural network regression}
\label{table_exp:1_2}
\end{table}

\begin{table}[ht!]
\centering
\begin{tabular}{c c} 
  \textbf{Settings} & \textbf{Value} \\ [0.5ex] 
 \hline
 ratio between dataset used for training C-MINE \\and computing estimated mutual information & 1:1 (use same dataset) \\ 
 number of samples in the dataset & 30000 \\
 clipping value $\tau$ for SMILE (clip log density ratio estimator for \\marginal dataset between $-\tau$ and $\tau$) & 1.0
 \\[1ex] 
\end{tabular}
\caption{training details for C-MINE and SMILE for Bayesian neural network regression}
\label{table_exp:1_3}
\end{table}

\bibliographystyle{IEEEtran}
\bibliography{ref}
\end{document}